\documentclass[12pt,upcase,hidelinks]{mitthesis}

\usepackage{amsmath, amsthm, amssymb}
\usepackage{microtype}
\usepackage{graphicx}
\graphicspath{{./images/}}
\usepackage{multirow}
\usepackage{rotating}
\usepackage{natbib}
\setcitestyle{numbers}
\usepackage{url}
\usepackage{booktabs}
\usepackage{hyperref}		% turns all internal references into hyperlinks
\usepackage{bm}
\usepackage{algorithmic}
\usepackage{chngpage}
\usepackage[resetlabels]{multibib}
\usepackage{comment}

\usepackage[]{pdfpages}

\usepackage{fancyhdr}
\newtheorem{proposition}{Proposition}
\newtheorem{definition}{Definition}

\newcites{New}{Publications}

\begin{document}
% Sherman 1
% Revision 1.1  92/04/22  13:08:20  epeisach

% BE SURE TO READ THE UNIVERSITY'S RULES ON WHAT FIELDS ARE REQUIRED OR ENCOURANGED FOR YOUR DEPARTMENT

% If your want to revise something for the first five page, please go to mitthesis.cls 
\title{Automatic Generation of Multiple-Choice Questions}

\author{Cheng Zhang}

% \date{September 2021}
\maketitle

\tableofcontents

% \newpage
% \listoffigures
% \newpage
% \listoftables

%\include{preface} 	% Preface is not part of the standard. I invented it for myself. Use it if you want by uncommenting.

	% Redefine the "plain" pagestyle to have pagenum in the header, right
	\fancypagestyle{plain}{
		\fancyhf{}
		\fancyhead[R]{\thepage}
	}

	\setcounter{page}{1}
	\pagenumbering{arabic}
	\pagestyle{plain}
	
\chapter{Introduction}

In an effort to build an online learning tool for helping students improve reading comprehension, it calls for a system to automatically generate adequate multiple-choice questions (MCQs) to assess student's understanding of a given article.
%'s main points. % from an arbitrary document. 
An article's main points include \textsl{direct} and \textsl{derived} points. 
A direct point is expressed in a declarative sentence. A derived point is inferred from multiple direct points, which could be a causal relation between them, an aggregation over them, or a conclusion drawn from them.
An MCQ typically consists of a question-answer pairs (QAPs) and a fixed number of distractors. 

% We study automatic generation of question-answer pairs (QAPs) with
Automatic generation of adequate MCQs can be divided into two tasks: (1) automatic generation of QAPs;  (2) automatic generation of distractors. Part of this dissertation is built on my prior work on QAP and distractor generation \cite{Zhang-Sun-Chen-Wang-2020,metaseq-2020,zhang2022downstream}.

\section{Automatic Question Generation}
We study automatic generation of QAPs with an emphasis on grammatical correctness of the questions and the suitability of the answers. 
By grammatical correctness we mean that the questions being generated are syntactically and semantically correct and conform to what a native speaker would say.
% By suitability we mean that the answers being generated consist of one correct answer and multiple adequate distractors.
% We refer to such QAPs as \textsl{adequate} QAPs.
% Other tasks of generating MCQs not addressed in this paper are how to provide adequate distractors for an answer.

Existing methods on QAP generation are based on handcrafted features or neural networks.
While they have met with certain success from different perspectives, the grand challenge of generating adequate QAPs still remains. 

%We present a new approach to tackling this challenge.
We present two approach to tackling this challenge from two different perspectives. 
(1) A deep-learning-based end-to-end Transformer with Preprocessing and Postprocessing Pipelines (TP3) for generating adequate QAPs, 
and (2) A sequence-learning-based MetaQG for generating adequate QAPs via meta sequence representations of sentences.

TP3 utilizes the power of Text-to-Text Transfer (T5) Transformer model.
It is a downstream task on a pretrained T5 that is finetuned on a QAP dataset, with a preprocessing pipeline to select appropriate answers and a postprocessing pipeline to filter undesirable questions.

While the TP3 system performs very well on unseen data and generates well-formed and grammatically correct questions from the given answer and context, it may still generate a few types of incoherent questions. For example, asking the type of verbs or asking clauses that express reason or purpose may generate incoherent QAPs. But these types of questions can be easily generated by MetaQA.
% -- a meta sequence learning based question generation system.

A meta sequence is a sequence of vectors comprising semantic and syntactic tags. 
In particular, we use a sequence of vectors to represent a sentence, where each vector consists of a semantic-role (SR) tag, a part-of-speech (POS) tag, and other syntactic and semantic tags, and we refer to such a sequence as a \textsl{meta sequence}.
We then present a scheme called MetaQA to learn meta sequences of declarative sentences and the corresponding interrogative sentences from a training dataset. % consisting of such sentences.
Combining and removing redundant meta sequences yields a set called MSDIP (Meta-Sequence-Declarative-Interrogative Pairs), with each element being a pair of an MD and corresponding MI(s), where MD and MI stand for, respectively, a meta sequence for a declarative sentence and for an interrogative sentence.
A trained MetaQA model generates QAPs for a given declarative sentence $s$ as follows:
Generate a meta sequence for $s$, find a best-matched MD from MSDIP, generates meta sequences for interrogative sentences according to the corresponding MIs and the meta sequence of $s$, identifies the meta-sequence answer to each MI, and coverts them back to text to form a QAP.

We implement MetaQA for the English language using SR, POS, and
NE (named-entity) tags.
%, and
%allow fuzzy representation when an existing tool fails to produce a tag for
%a given word. 
We then train MetaQA using a moderate initial dataset and show that MetaQA generates efficiently a large number of adequate QAPs with an accuracy of 97\% on the official SAT practice reading tests.
These tests contain a large number of declarative sentences in different patterns, and there is no match in the initial MSDIP for some of these sentences. After learning interrogative for some of these sentences, MetaQA successfully generate many more adequate QAPs.

\section{Automatic Distractor Generation}
Given a QAP for a given article, we also study how to generate adequate distractors that are grammatically correct and semantically related to the correct answer in the sense that the distractors, while incorrect, look similar to the correct answer with a sufficient distracting effect---that is, it should be hard to distinguish distractors from the correct answer without some understanding of the underlying article.
An distractor could be a single word, a phrase, a sentence segment, or a complete sentence.
It must satisfy the following requirements:
(1) it is an incorrect answer to the question;
(2) it is grammatically correct and consistent with the underlying article;
(3) it is semantically related to the correct answer; and
(4) it provides distraction so that the correct answer could be identified only with some understanding of the underlying article.

In particular, we are to generate three adequate distractors for a QAP to form an MCQ. One way to generate a distractor is to substitute a word or a phrase contained in the answer with an appropriate word or a phrase that maintains the original part of speech. 
Such a word or phrase could be an answer itself or contained in an answer sentence or sentence segment. For convenience, we refer to such a word or phrase as a target word. 

If a target word is a number with an explicit or implicit quantifier,
%is a point in time, a numerical value, 
or anything that can be converted to a number, we call it a type-1 target.
If a target word is a person, location, or organization, we call it a type-2 target. 
Other target words (nouns, phrasal nouns, adjectives, verbs, and adjectives) are referred to as type-3 targets.
We use different methods to generate distractors for targets of different types.

Our distractor generation method is a combination of part-of-speech (POS) tagging \cite{toutanova2003feature}, 
named-entity (NE) tagging \cite{nadeau2007survey,ali2010automation,peters2017semi}, 
semantic-role labeling \cite{martha2005proposition,shi2019simple},
regular expressions, domain knowledge bases on people, locations, and organizations,
word embeddings (such as Word2vec \cite{10.5555/2999792.2999959}, GloVe \cite{pennington-etal-2014-glove}, Subwords \cite{bojanowski2017enriching}, and spherical text embedding \cite{sphericalembedding2019}), word edit distance \cite{levenshtein1966binary}, WordNet (https://wordnet.princeton.edu), and some other algorithms.
We show that, via experiments, our method can generate adequate distractors for a QAP to form an MCQ with a high successful rate.  %Introduction
\chapter{Related Work}

\section{Automatic Question Generation}
Automatic question generation (QG), first studied by Wolfe \cite{wolfe1976automatic} as a means to aid independent study, has since attracted increasing attentions in two lines of methodologies: transformative and generative. 

\subsection{Transformative Methods}
Transformative methods transform key phrases from a given single declarative sentence into factual questions.
Existing methods are rule-based on syntax, semantics, or templates.

Syntactic-based methods follow the same basic strategy: Parse sentences using a syntactic parser to identify key phrases and transform a sentence to a question based on syntactic rules. 
These include methods to identify key phrases from input sentences and use syntactic rules for different types of questions \cite{varga2010wlv}, generate questions and answers using a syntactic parser, a POS tagger, and an NE analyzer \cite{ali2010automation}, transform a sentence into a set of questions using a series of domain-independent rules \cite{danon2017syntactic}, and generate questions using relative pronouns and adverbs from complex English sentences \cite{khullar2018automatic}. 

Semantic-based methods create questions using predicate-argument structures and semantic roles \cite{mannem2010question}, semantic pattern recognition  \cite{mazidi2014linguistic}, subtopics based on Latent Dirichlet Allocation \cite{chali2015towards}, or
%generate factual questions using 
semantic-role labeling %as the main form of text analysis 
\cite{flor2018semantic}.

These methods are similar. The only difference is that semantic-based methods use semantic parsing while syntactic-based methods use syntactic parsing to determine which specific words or phrases should be asked. In a language with many syntactic and semantic exceptions, such as English, these methods would require substantial manual labor to construct rules.

Template-based methods are for special-purpose applications with built-in templates. Research in this line devises a Natural Language Generation Markup Language (NLGML) \cite{cai2006nlgml}; 
uses a phrase structure parser to parse text and construct questions using enhanced XML \cite{rus2007experiments};
devise a self-questioning strategy to help children generate questions from narrative fiction \cite{mostow2009generating};
use informational text to enhance the self-questioning strategy \cite{chen2009aist};
apply pattern matching, variables, and templates to transform source sentences into questions similar to NLGML \cite{wyse2009generating};
defines a question template as pre-defined text with placeholder variables to be replaced with content from the source text \cite{lindberg2013automatic}; 
or incorporates semantic-based methods into a template-based method to support online learning \cite{lindberg2013generating}.

\subsection{Generative Methods}

Recent advances of neural-network research provide new tools to build generative models.
For example, the attention mechanism can help determine what content in a sentence should be asked \cite{luong-etal-2015-effective}, and the sequence-to-sequence  \cite{bahdanau2014neural,cho-etal-2014-learning} and the long short-term memory  \cite{Sak2014LongSM}  mechanisms are used to generate words to form a question (see, e.g., \cite{du-etal-2017-learning,duan-etal-2017-question,Harrison_2018,sachan-xing-2018-self}).
These models generate questions without the corresponding correct answers. 
%Moreover, training these models require a dataset comprising over 100K questions.
To address this issue, % the problem of generating questions without answers, 
researchers have explored ways to encode a passage (a sentence or multiple sentences) and an answer word (or a phrase) as input, and determine what questions are to be generated for a given answer \cite{10.1007/978-3-319-73618-1_56,zhao-etal-2018-paragraph,song-etal-2018-leveraging}. 
However, as pointed out by
Kim et al. \cite{Kim_2019}, these methods could generate answer-revealing questions, namely, questions contain in them the corresponding answers. They then devised a new method by encoding answers separately, at the expense of learning substantially more parameters. 

More recently, researchers have explored how to use pretrained transformers to generate answer-aware questions \cite{dong2019unified,Zhang2019AddressingSD,Zhou2019QuestiontypeDQ,qi-etal-2020-prophetnet,su-etal-2020-multi,10.1145/3442381.3449892}.
For example, Kettip et al. \cite{Kriangchaivech190905017} presented an architecture for a transformer to generate questions. Rather than fully encoding the context and answers as they appear in the dataset, they applied certain transformations such as the change of named entities both on the context and the answer. 
Lopez et al. \cite{Lopez2020TransformerbasedEQ} finetuned the pretrained GPT-2 \cite{radford2019language} transformer without using any additional complex components or features to enhance its performance.
Chen \cite{Chen2020ReinforcementLB} described a fully transformer-based reinforcement learning generator evaluator architecture to generate questions.

The recent introduction of T5 has escalated NLP research in a number of ways.
%, which achieves the state-of-the-art results on several NLP task trained on a newer and larger text corpus. % called C4
%\subsection{An Overview of T5}
%
%We present a deep-learning-based end-to-end question generation model to utilize the power of Text-to-Text Transfer Transformer (T5) \cite{raffel2020exploring}. 
T5 is a encoder-decoder text-to-text transformer using the teacher forcing method on a wide variety of NLP tasks, including text classification, question answering, machine translation, and abstractive summarization. Unlike other transformer models (e.g. GPT-2 \cite{radford2019language}) that take in text data after converting them to corresponding numerical embeddings, T5 handles each task by taking in data in the form of text and producing text outputs. 
% Fig \ref{fig:1} depicts the architecture of T5 \cite{raffel2020exploring}. 
%The model has an encoder-decoder based transformer architecture \cite{raffel2020exploring}. 

% \begin{figure}[h]
%   \centering
%   \includegraphics[width= \linewidth]{T5Architecture}
%   \caption{T5 Architecture}
%   \label{fig:1}
% %   \Description{}
% \end{figure}

% In particular,
% the encoder consists of a stack of identical layers. Every layer is composed of two sub-layers. The first sub-layer of each encoder layer is a multi-head self-attention mechanism. The second sub-layer  is a fully connected feed-forward network. Residual connections are employed around these sub-layers, each followed by a normalization layer. 
% %
% The decoder also consists of a stack of identical layers. In addition to the two sub-layers already present in the encoder layer, the third sub-layer performs multi-head attention on the output received by the encoder stack. Residual connections are employed around these sub-layers, such as those in the encoder, each followed by a normalization layer. 

%
%
%most recent innovation on question generation train a %pretrained %the Text-to-Text Transfer Transformer (
%T5 model \cite{raffel2020exploring} %, it pretrained upon 
%on a newer and larger text corpus called C4, which achieves the state-of-the-art results on several NLP tasks.
Taking the advantage of pretrained T5, 
Lidiya et al. \cite{mixqg.2110.08175} combined nine question-answering datasets to finetune a single T5 model and evaluated generated questions using a new semantic measure called BERTScore \cite{bert-score}.
Their method achieves so far the best results. % state-of-art result.
We present a finetuned T5 model on a
single SQuAD dataset %with preprocessing and postprocessing pipelines 
to produce better results. %than the Lidiya's approach.

\section{Automatic Distractor Generation}
Methods of generating adequate distractors for MCQs are typically following two directions: domain-specific knowledge bases and semantic similarity \cite{pho-etal-2014-multiple,CH2018Automatic}.

Methods in the first direction are focused on lexical information 
% synonym, hyponym and hypernym
have used part-of-speech(POS) tags, word frequency, WordNet, domain ontology, distributional hypothesis, and pattern matching, to find the target word's synonym, hyponym and hypernym as distractor candidates. \cite{10.3115/1118894.1118897,10.1007/978-3-642-28885-2_19,Susanti2015AutomaticGO}
% Liu et al \cite{liu-etal-2005-applications} Collecting similar candidates of the target word in terms of their frequency and dictionary-based collocation.

Methods in the second direction %semantic similarity methods 
analyze the semantic similarity of the target word using Word2Vec model for generating distractor candidates \cite{Jiang2017DistractorGF,Susanti2018AutomaticDG}. 

However, it is difficult to use Word2Vec or other word-embedding methods to find adequate distractors for polysemous answer words.
Moreover, previous efforts have focused on finding some forms of distractors, instead of making them look more distracting. This paper is an attempt to tackle these issues. %Related Work
\chapter{Deep-learning-based Question Generation}

\section{Description of TP3}\label{sec:finetuning}

We describe how we train and finetune a pretrained T5 transformer for our downstream task of question generation and use a combination of various NLP tools and algorithms to build the preprocessing and postprocessing pipelines for generating QAPs. %the QG system.

%More details can be referred from the original paper \cite{raffel2020exploring}.

%\subsection{Selecting a dataset to finetune T5}

%To use the T5 model for our downstream task of question generation, we finetune the T5 model on QAP datasets.

There are a number of public QAP datasets available for fine-tuning T5, including RACE \cite{lai2017large}, CoQA \cite{reddy2019coqa}, and SQuAD \cite{Rajpurkar_2016}. 
RACE is a large-scale dataset collected from Gaokao English examinations over the years, where Gaokao is the national college entrance examinations held once every year in mainland China. It consists of
 more than 28,000 passages and nearly 100,000 questions, including cloze questions.
CoQA is a conversational-style question-answer dataset. It contains a series of interconnected questions and answers %that appear 
in conversations. % about a text. 
SQuAD is a reading comprehension dataset, consisting of more than 100,000 QAPs %and corresponding answers 
posted by crowdworkers on a set of Wikipedia articles.

Among these datasets, %Based on question types, and the 
SQuAD is more commonly used in the question generation research. We use SQuAD to finetune pretrained T5 models. 
%
%The T5 model is tasked with generating a relevant question for a given answer and a context for that answer.
For each QAP and corresponding context extracted from the SQuAD training dataset, we concatenate the answer and the context with markings in the format of $\langle answer\rangle answer\_text \langle context\rangle context\_text$ as input, with the question as the target, where the context is the entire article for the QAP in SQuAD. 
%for calculating loss during training, 
We then set the maximum input length to 512 and the target length to 128 to avoid infinite loops and repetitions of target outputs. We feed the concatenated text input and question target into a pretrained T5 model for fine-tuning and use AdamW \cite{loshchilov2018decoupled} as an optimizer with various learning rates to obtain a better model.

%We discuss the experiment results in Section \ref{sec:evaluations}
%large auto lr: 0.0019054607179632484
%base auto lr: 0.0004365158322401656
%We'd like to explore a large number of different learning rates. 
To explore various learning rates, we first use automatic evaluation methods to narrow down a smaller range of the learning rates and then use human judges to determine the best learning rate.
In particular, we first finetune the base model with a learning rate of $1.905 \times 10^{-3}$ and the large model with a learning rate of $4.365 \times 10^{-4}$. The learning rates are calculated %selected 
using the Cyclical Learning Rates (CLR) method \cite{7926641}, which is used to find automatically the best global learning rate. % and schedule for the global learning rates.
%, and it is used in many well-known deep learning frameworks, such as PyTorch and TensorFlow.
Evaluated by human judges, we found that %the generated  are fair but not ideal as our expected. We also noticed that 
the best learning rate calculated by CLR is always larger than the actual best learning rate in our experiments.

We then finetune T5-Base and T5-Large with dynamic learning rates from the learning rate calculated by CLR with a reduced learning rate for each epoch. For example, we finetune T5-Base starting from a learning rate of $1.905 \times 10^{-3}$ and multiply the previous learning rate by 0.5 for the current epoch until % multiplication in each epoch decreased to 
the learning rate of $1.86 \times 10^{-6}$ is reached. Likewise, we finetune T5-Large in the same way starting from $4.365 \times 10^{-4}$ until
%,  with 0.5 multiplication in each epoch decreased to 
the learning rate of $1.364 \times 10^{-5}$ is reached. However, the generated results are still below expectations. 

We therefore proceed to 
%Since the learning rate calculated by CLR is not reliable, we then tried to 
finetune the models with various learning rates we choose. In particular,
%According to previous models training experience, 
we first finetune T5-Base with a constant learning rate from $10^{-3}$ to $10^{-4}$ with a $2.5 \times 10^{-4}$ decrement for each model, and from $10^{-4}$ to $10^{-5}$ with a $2.5 \times 10^{-5}$ decrement for each epoch. Likewise, we  
finetune T5-Large with a learning rate from $10^{-4}$ to $10^{-5}$ with a $2 \times 10^{-5}$ decrement for each epoch, and from $10^{-5}$ to $10^{-6}$ with a $2 \times 10^{-6}$ decrement for each epoch.

%The following metrics are used to automatically evaluate the performance of question generation from the T5 models:
%
%BLEU scores measure the quality of text that has been translated by a machine from one natural language to another using n-grams. We used a cumulative 4-gram BLEU score (B4) as an evaluation metric.
%
%ROUGE-L uses statistics based on the Longest Common Subsequence (LCS) to evaluate recall by how many words in reference sentences are used in predicted sentences.
%
%METEOR is a precision-based metric for evaluating machine-translation out- put.
%
%BERTScore is a contextual embedding based metric for ...

Evaluated using BLEU \cite{papineni-etal-2002-bleu}, ROUGE \cite{lin-2004-rouge}, METEOR \cite{banerjee-lavie-2005-meteor} and BERTScore \cite{bert-score}, we find that the learning rates ranging from $10^{-4}$ to $10^{-5}$ for T5-Base and the learning rates ranging from $10^{-5}$ to $10^{-6}$ for T5-Large perform better.
%large model performed better results.
%
%After human evaluation, 
%Specifically, 
%we find that T5-Base with a learning rate of $3 \times 10^{-5}$ and T5-Large with a learning rate of $6 \times 10^{-6}$ produce the best results, and so we use these models to generate questions. 
Moreover, as expected,
the overall performance of T5-Large is better than T5-Base. 

Tables \ref{tab:auto-evaluation-base} and \ref{tab:auto-evaluation-large}
depict the measurement results for T5-Base and T5-Large, respectively, where R1, R2, RL, and RLsum stand for, respectively,
Rouge-1, Rouge-2, Rouge-L, and Rouge-Lsum. %The average is taken over all measures after multiplying the BERTScore scores by 100. 
The boldfaced number indicates the best in its column. It is evident that T5-Base with the
learning rate of $3 \times 10^{-5}$ and
T5-Large with the learning rate of $8 \times 10^{-6}$ produce the best results.
For convenience, we refer to these two finetuned models as T5-Base-SQuAD$_1$ and T5-Large-SQuAD$_1$
to distinguish them with the existing T5-Base-SQuAD model.
We sometimes also denote T5-Base-SQuAD$_1$ as T5-SQUAD$_1$ when there is no
confusion of what size of the dataset is used to pretrain T5.
% and so we use it to generate questions. 
%picked these models for our QG system.

\begin{table}[h]
\footnotesize
\centering
\caption{Automatic Evaluation of T5-Base-SQuAD$_1$}
\begin{tabular}{l|c|c|c|c|c|c|c|c}
\hline
\textbf{Learning Rate} & \textbf{BLEU}  & \textbf{R1}    & \textbf{R2}    & \textbf{RL}    & \textbf{RLsum} & \textbf{METEOR} & \textbf{BERTScore} & \textbf{Average} \\ \hline
5e-5              & 20.01 & 50.71 & 28.38 & 46.59 & 46.61 & 45.46  & 51.51 & 41.32     \\ \hline
3e-5              & \textbf{22.63} & \textbf{54.90} & \textbf{32.22} & \textbf{50.97} & \textbf{50.99} & \textbf{48.98}  & \textbf{55.82} & \textbf{45.22}     \\ \hline
2.5e-5            & 22.50 & 54.36 & 31.93 & 50.49 & 50.50 & 48.64  & 55.61 & 44.86     \\ \hline
1e-5              & 20.17 & 50.46 & 28.38 & 46.79 & 46.81 & 44.97  & 51.82 & 41.34     \\ \hline
Dynamic %1.2e-4 $\rightarrow$ 3e-5 
& 20.57 & 51.88 & 28.99 & 47.67 & 47.68 & 47.38  & 53.34 & 42.50     \\ \hline
%Dynamic automatic      & 4.74  & 22.49 & 7.54  & 19.04 & 19.05 & 19.84  & 13.94 & 15.23     \\ \hline
\end{tabular}
\label{tab:auto-evaluation-base}
\end{table}

\begin{table}[h]
\footnotesize
\centering
\caption{Automatic Evaluation on T5-Large-SQuAD$_1$}
\begin{tabular}{l|c|c|c|c|c|c|c|c}
\hline
\textbf{Learning Rate} & \textbf{BLEU}  & \textbf{R1}    & \textbf{R2}    & \textbf{RL}    & \textbf{RLsum} & \textbf{METEOR} & \textbf{BERTScore} & \textbf{Average} \\ \hline
3e-5                    & 23.01 & 54.49 & 31.92 & 50.51 & 50.51 & 50.00  & 56.19 & 45.23    \\ \hline
1e-5                    & 23.66 & 51.88 & 32.88 & 51.43 & 51.42 & 50.53  & 56.65 & 45.50   \\ \hline
8e-6                    & 23.83 & \textbf{55.48} & \textbf{33.08} & \textbf{51.58} & \textbf{51.58} & 50.61  & \textbf{56.94} & \textbf{46.15}   \\ \hline
6e-6                    & \textbf{23.84} & 55.24 & 32.91 & 51.35 & 51.35 & \textbf{50.70}  & 56.57 & 45.99   \\ \hline
Dynamic                 & 20.86 & 52.00 & 29.46 & 48.03 & 48.03 & 47.68  & 53.85 & 42.84   \\ \hline
\end{tabular}
\label{tab:auto-evaluation-large}
\end{table}

\section{Processing Pipelines} %Candidate Answers selecting and filtering}
\label{sec:P3}
%preprocessing

\begin{figure}[h]
  \centering
  \includegraphics[width= \linewidth]{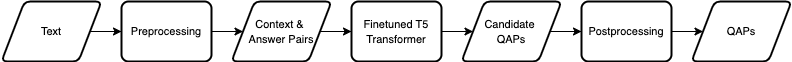}
  \caption{TP3 Architecture}
  \label{fig:2}
%   \Description{}
\end{figure}

The processing pipelines consist of preprocessing to select appropriate answers, question generation, and postprocessing to filter undesirable questions (see Fig. \ref{fig:2}).

\subsection{Preprocessing}

We observe that how to choose an answer would affect the quality of a question generated for the answer.
%however, not all answers can generate a satisfying question, 
%In general, T5 models work well on unseen data. and generates well-formed and grammatically correct questions from the given answer and context. There are, however, a few types of questions that could lead to generating incoherent questions. For example, asking about the type of verbs or asking about the clauses that express reasons or purposes may generate incoherent QAPs. 
We use a combination of NLP tools and algorithms to construct a preprocessing pipeline for
selecting appropriate answers as follows: % and filter undesirable questions. 
%
%In preprocessing stage, we select candidate answers by following 3 steps:
\begin{enumerate}
\item \textit{Remove unsuitable sentences}. We
first remove all interrogative and imperative sentences from the given article. We may do so 
by, for instance, simply removing any sentence that begins with a WH word or a verb and any sentence that ends with a question mark.
We then use semantic-role labeling \cite{Shi2019SimpleBM} to analyze sentences and remove those
that do not have any one of the following semantic-role tags: %ach sentence must consist of 3 semantic roles: 
subject, verb, and object. % we filtered out if any sentence doesn't satisfy this requirement. 
For each remaining sentence, if the total number of words contained in it, excluding stop words, is less than 4, then remove this sentence.
We then label the remaining sentences as \textit{suitable} sentences.

\item \textit{Remove candidate answers with inappropriate semantic-role labels}.
%The basic idea of candidate answers extraction is that, 
%We'd like to extract 
Nouns and phrasal nouns are candidate answers. But not any noun or phrasal noun would be suitable to be an answer.
We'd want a candidate answer to associate with a specific meaning. 
%In particular, we'd select candidate answer
%with a semantic-role label of subject, object, or manner. 
%To do so, 
%we first remove candidate answers with inappropriate semantic-role labels.
%We'd also like to extract 
%semantic roles with nouns. % as the main word in the role.
%To achieve this, 
Specifically, if a noun in a suitable sentence is identified as a name entity \cite{Peters2017SemisupervisedST}
or has a semantic-role label in the set of $\{$ARG, TMP, LOC, MNR, CAU, DIR$\}$, then keep it
as a candidate answer and remove the rest,
where ARG represents subject or object,
TMP represents time, LOC represents location, MNR represents manner, CAU represents cause, and DIR represents direction.
%
%with the following semantic-role labels as candidate answers and remove: ARG, TMP, LOC, MNR, and CAU as candidate answers, as well as words with named-entity labels \cite{Peters2017SemisupervisedST}. 
If a few candidate nouns occur consecutively, we treat the sequence of these nouns as a candidate answer phrase.

For example, in the sentence ``The engineers at the Massachusetts Institute of Technology (MIT) have taken it a step further changing the actual composition of plants in order to get them to perform diverse, even unusual functions", the phrase ``Massachusetts Institute of Technology" is recognized as a named entity, without a semantic-role label. Thus, it should not be selected as an answer. If it is selected, then the following QAP
(``Where is MIT located", ``Massachusetts Institute of Technology") will be generated, which is inadequate. 

\item \textit{Remove or prune answers with inadequate POS tags}.
Using semantic-role labels to identify what nouns to keep does not always work.
%works well most of the times, but
%there are situations that it would produce inadequate QAPs. 
%words are labeled wrong or labels redundant words. 
For example, the phrasal noun ``This widget" in the sentence "This widget is more technologically advanced now"
has a semantic-role label of ARG1 (subject), which leads to the generation of the following question: 
``What widget is more technologically advanced now?" It is evident that this QAP is inadequate even though it is
grammatically correct. 
Note that ``This" has
a POS (part-of-speech) tag of PDT (predeterminer).
%This example indicates that semantic-role labeling alone is insufficient and POS tagging should also be used.
%is not necessary for the answer and it is labeled as predeterminer (PDT) by POS tagging, if the model generates question according to this answer phrase, it may generate inappropriate question "What widget is more technologically advanced now?" with answer "This widget".
For another example, while the word ``now" in the sentence 
has a semantic-role label of TMP (time), its POS tag is RB (adverb).
In general, we remove nouns with a POS tag in $\{$RB, RP, CC, DT, IN, MD, PDT, PRP, WP, WDT, WRB$\}$ 
or prune words with such a POS tag at either end of a phrasal noun.
%
%should be removed. In general,
%, it should not be used as answer for question generation.
%we remove such answers or keep the noun component in a phrase by pruning the words at either end 
%with %in an answer phrase according to 
%the following POS tags: 
%RB, RP, CC, DT, IN, MD, PDT, PRP, WP, WDT, WRB. 
After this treatment, the candidate answer ``now" is removed and the candidate answer phrase ``This widget" is pruned to ``widget". For this answer and the input sentence, the following question is generated: ``What is more technologically advanced now?" Evidently this question is more adequate.

\item \textit{Remove common answers}. %%%needs work
%Through experiments, % and generalization, 
We observe that certain candidate answers, such as ``anyone", ``people", and ``stuff", 
would often lead to generation of inadequate questions. 
%For example, "anyone", "people", "stuff" etc. 
Such words tend to be common words that should be removed. 
We do so by looking up the probabilities of 1-grams from the
%language model of the 
Google Books Ngram Dataset \cite{doi:10.1126/science.1199644}. If the probability of a noun word is greater 
than 0.15\%, we remove its candidacy. Likewise, we may also treat noun phrases by looking up the probabilities $n$-grams for $n > 1$, but doing so would incur much more processing time.

\item \textit{Filtering answers appearing in clauses}.
We observe that a candidate answer appearing in the latter part of a clause would often lead to a generation of an
inadequate QAP. Such candidate answers would appear at lower levels in a dependency tree. We use the following procedure to identify such candidate answers: For each remaining sentence $s$, we first generate its dependency tree \cite{varga2010wlv}. Let $h_s$ be the height of the tree. Suppose that a candidate answer $a$ appears in a clause contained in $s$. If $a$ is a single noun, let its height in the tree be $h_a$. If $a$ is a phrasal noun, 
let the average height of the heights of the words contained in $a$ be $h_a$. If $h_a \geq \tfrac{2}{3}h_s$, then remove $a$.

Take the following sentence as an example: ``While I tend to buy a lot of books, these three were given to me as gifts, which might add to the meaning I attach to them." In 
this sentences, the following noun ``gifts” and phrasal nouns ``a lot of books" and ``the meaning I attach to them"are labeled as object.
However, T5
% is labeled as object, where 
%``a lot of books" and ``gifts" both reference the same semantic-role label of object and the underlying transformer may not be able to
resolves multiple objects poorly, 
and if we choose ``the meaning I attach to them" as an answer, T5 will generate the following question: 
%answer phrase "the meaning I attach to them" is labeled as object in " which" clause in the sentence , since  and the model sometime can't resolve multiple objects appropriately, the phrase may leads to generate an inadequate question 
``What did the gifts add to the books", which is inadequate. Since this phrasal noun appears in a clause and at a lower level of the dependency tree, it is removed from being selected as a candidate answer.

%and its height is at least 2/3 of the height of the tree, then remove it.
%is higher than that of the verb node, 
%and if it higher than 75\% of the dependency tree height, we filter the it out, 
%If the answer is a phrasal noun, we take the average height of each word in the phrase as its height. 

\item \textit{Removing redundant answers}.
%We remove duplicate answers.
% by Named Entity labeling and semantic-role labeling, 
%If a candidate answer is contained in a longer answer phrase, we remove the shorter one.
If a candidate answer word or phrase is contained in another candidate answer phrase and appear in the same sentence,
we extract from the dependency tree of the sentence the subtree $T_s$ for the shorter candidate phrase and subtree $T_l$ for the longer candidate phrase, then $T_s$ is also a subtree of $T_l$. If $T_s$ and $T_l$ share the same root, %is same as the root part of the $h_l$, we say 
then the shorter candidate answer is more syntactically important than the longer one, and so we remove the longer candidate answer. Otherwise, remove the shorter candidate answer.
 
Take the sentence ``The longest track and field event at the Summer Olympics is the 50-kilometer race walk, which is about five miles longer than the marathon" as an example.  
The shorter phrase ``Summer Olympics" is recognized as a named entity, 
%and a longer phrase "The longest track and field event at the Summer Olympics" is labeled as subject by semantic role.
which leads to the generation of the following inadequate QAP: (``What is the longest track and field event", ``Summer Olympics). 
On the other hand,
the longer phrase ``The longest track and field event at the Summer Olympics" is labeled as subject for its semantic role, which leads to the generation of the following adequate QAP: (``What is the 50-kilometer race walk", ``The longest track and field event at the Summer Olympics").
Since the root word for the longer phrase is ``event" that is not contained in the shorter phrase, so the shorter phrase is removed to avoid generating the inadequate QAP.

\end{enumerate}

\subsection{Question generation}

After extracting all candidate answers from the preprocessing pipeline, for each answer extracted, we use three adjacent sentences as the context, with the middle sentence containing the answer, and concatenate the answer and the context with marks into the following format as input to a fine-turned T5 model: $\langle answer\rangle$answer\_text$\langle context\rangle$context\_text, to generate candidate questions.
We note that the greedy search in the decoder of the T5 model does not guarantee the optimal result, we use beam search with 3 beams to select the word sequences with the top 3 probabilities from the probability distribution and acquire 3 candidate questions.
We then concatenate each candidate question with the corresponding answer as a new sentence and generate
an embedding vector representation for it using %map the sentence to %dense vector space 
%sentence embeddings by utilizing 
the pretrained  RoBERTa-Large model \cite{Liu2019RoBERTaAR,reimers-2019-sentence-bert}, and select the most semantically similar question to the context as the final target question.

\subsection{Postprocessing}

Recall that in the preprocessing pipeline, %the answers selecting and filtering stage, 
we have removed inappropriate candidate answers.
% that would lead to generating appropriate questions.
However, some of the remaining answers may still lead to generating inappropriate questions. 
Thus, in the postprocessing pipeline, we proceed to remove inadequate questions as follows:
%by following 3 steps to tackle this problem:
\begin{enumerate}
\item \textit{Remove questions that contain the answers}.
Remove a question if the corresponding answer or the main body of the answer is contained in the question.
%\hl{Changes start here}
If the answer includes a clause, we extract the main body of the answer as follows:
Parse the answer to constituency tree \cite{Joshi2018ExtendingAP} and remove the subtree rooted with a subordinate clause label SBAR, the remaining part of the phrase is the main body of the answer.

For example, in the sentence ``The first, which I take to reading every spring is Ernest Hemningway's A Moveable Feast", ``The first, which I take to reading every spring" is labeled as subject. Using it as a candidate answer  generates an inadequate question for the answer ``What is the first book I reread?" Note that the phrase ``The first" can be extracted as the main body of the answer, which is contained in the question. Thus, this QAP is removed.
%the generated question can be removed.
%\hl{Changes end here}

\item \textit{Remove short questions}.
If the generated question, after removing stop words, consists of only one word, then remove the question.
%we filter this question. 
For example, ``What is it?" and ``Who is she?" will be removed because after removing stop words,
the former becomes ``What" and the latter becomes ``Who". On the other hand, ``Where is Boston?" will remain.

\item \textit{Remove unsuitable questions}.
Recall that we generate the question from the adjacent three sentences in the article, with the middle sentence containing the answer. However, the middle sentence may not be the only sentence containing the answer. In other words,
the first or the last sentences may also contain the answer. 
Assuming that all three sentences contain the answer, our finetuned T5 transformer may generate a question based on the first sentence or the last sentence. If the first sentence or the last sentence is not a suitable sentence
we labeled in the preprocessing pipeline, 
%our generatable sentence list which we described in the preprocessing stage, 
the question being generated may be in appropriate. 
We'd want to make sure that the question is generated for a suitable sentence.
For this purpose, we first identify which sentence the question is generated for. In particular,
%it may cause generating inappropriate questions.
let $s_i$ for $i=1,2,3$ be the 3 sentences and $(q,a)$ be the question generated for answer $a$.
Let $QA$ denote the union of the set of words in $q$ and the set of words in $a$.
Likewise, let $S_i$ be the set of words in $s_i$. If $QA \cap S_i$ is the largest among the other two
intersections, then $q$ is likely generated from $s_i$ for $a$. If $s_i$ is not suitable, then
remove $q$.

Note that we may also consider word sequences in addition to word sets. For example, we may consider longest common subsequences or longest common substrings when comparing two word sequences. But in our experiments, they don't seem to 
add extra benefits.
%a word sequence that concatenate the word sequence in the generated question 
%and the word sequence in the answer, and form three other word sequences each of which corresponds to an input sentence.
%
%in the order they appear and the answer into a word sequence, and place words in each sentence in the order they appear into another word sequence.
%We then count the number of words in the intersection of two word sequences and the number of longest common subsequence for question answer with each sentence. If the sentence has more intersection and more longest common subsequence with the question and answer, we think the question is generated from that sentence, and if the sentence is not in the generatable sentence list, we filter out the corresponding generated question.

%(4) Prune question,  parser question to cp tree,  if question contains clause (tree node contains SBAR), and the question longer than 10 words, remove the leaves followed by when, where, which. 
\end{enumerate}

\section{Running Samples}
Suppose that the following sentences are given as a article:

Returning to a book you've read many times can feel like drinks with an old friend. There's a welcome familiarity - but also sometimes a slight suspicion that time has changed you both, and thus the relationship. But books don't change, people do. And that's what makes the act of rereading so rich and transformative.
The beauty of rereading lies in the idea that our bond with the work is based on our present mental register. It's true, the older I get, the more I feel time has wings. But with reading, it's all about the present. It's about the now and what one contributes to the now, because reading is a give and take between author and reader. Each has to pull their own weight.
There are three books I reread annually. The first, which I take to reading every spring is Ernest Hemningway's A Moveable Feast. Published in 1964, it's his classic memoir of 1920s Paris. The language is almost intoxicating, an aging writer looking back on an ambitious yet simpler time. Another is Annie Dillard's Holy the Firm, her poetic 1975 ramble about everything and nothing. The third book is Julio Cortazar's Save Twilight: Selected Poems, because poetry. And because Cortazar. 
While I tend to buy a lot of books, these three were given to me as gifts, which might add to the meaning I attach to them. But I imagine that, while money is indeed wonderful and necessary, rereading an author's work is the highest currency a reader can pay them. The best books are the ones that open further as time passes. But remember, it's you that has to grow and read and reread in order to better understand your friends.

The following QAPs are generated by TP3:

\begin{enumerate}
\item Question: Who wrote Holy the Firm?

Answer: Annie Dillard

\item Question: Who wrote A Moveable Feast?

Answer: Ernest Hemningway

\item Question: What is the first book I reread every spring?

Answer: Ernest Hemningway's A Moveable Feast

\item Question: Which book by Annie Dillard is a 1975 ramble about everything and nothing?

Answer: Holy the Firm

\item Question: What is the name of the book I reread every year?

Answer: Julio Cortazar's Save Twilight

\item Question: What Ernest Hemingway book do I reread every spring?

Answer:  A Moveable Feast

\item Question: What is intoxicating about Ernest Hemingway's A Moveable Feast?

Answer: The language

\item Question: How many books do I tend to buy?

Answer: a lot of books

\item Question: Who is Ernest Hemningway?

Answer: an aging writer

\item Question: Rereading an author's work is what do I imagine a reader can pay them?

Answer: the highest currency

\end{enumerate}

\section{Evaluations}\label{sec:evaluations}

To evaluate the quality of QAPs generated by TP3-Base and TP3-Large, we use the standard automatic evaluation metrics as well as
human judgments.

\subsection{Automatic evaluations}

We first compare T5-SQuAD$_1$ with the exiting QG models 
with the standard automatic evaluation metrics as before:
BLEU, % \cite{10.3115/1073083.1073135}, 
ROUGE-1 (R1), ROUGE-2 (R2), ROUGE-L (RL), ROUGE-LSum (RLsum), % \cite{lin-2004-rouge}, 
METEOR (MTR), % \cite{banerjee-lavie-2005-meteor}) to automatically evaluate our finetuned models, those n-grams-based metrics was used to evaluate the syntactic reconstruction ability of the models. Besides the n-grams-based metrics, we also used BERTScore \cite{bert-score} as semantic-based metric to evaluate the semantic reconstruction ability of the models.
and BERTScore (BScore). 
Since most existing QG models are based on pretrained transformers with the base dataset, we will
compare T5-Base-SQuAD$_1$ with the existing QG models. 
%The results are shown on
%Table \ref{tab:auto-evaluation}.

\begin{table}[h]
\footnotesize
\centering
\caption{Automatic evaluation results}
\begin{tabular}{l|c|c|c|c|c|c|c|c|c}
\hline
\textbf{Model}  & \textbf{Size}  & \textbf{BLEU} & \textbf{R1}  & \textbf{R2} & \textbf{RL} & \textbf{RLsum}          & \textbf{MTR}    & \textbf{BScore}     &\textbf{Average}  \\ \hline
ProphetNet  & Large & 22.88          & 51.37          & 29.48          & 47.11          & 47.09          & 41.46          & 49.31     & 41.24   \\ \hline
BART-hl     & Base  & 21.13          & 51.88          & 29.43          & 48.00          & 48.01          & 40.23          & 54.33     & 41.86    \\ \hline
BART-SQuAD  & Base  & 22.09          & 52.75          & 30.56          & 48.79          & 48.78          & 41.39          & 54.86     & 42.75    \\ \hline
T5-hl       & Base  & 23.19          & 53.52          & 31.22          & 49.40          & 49.40          & 42.68          & 55.48     & 43.56    \\ \hline
T5-SQuAD    & Base  & \textbf{23.74} & 54.12          & 31.84          & 49.82          & 49.81          & 43.63          & 55.68     & 44.09    \\ \hline
MixQG$_1$       & Base  & 23.53          & 54.39          & 32.06          & 50.05          & 50.02          & 43.83          & 55.66     & 44.22    \\ \hline
MixQG$_2$       & Base  & 23.74          & 54.28          & 32.23          & 50.35          & 50.34          & 43.91          & 55.71     & 44.37    \\ \hline
MixQG-SQuAD & Base  & 23.46          & 54.48          & 32.18          & 50.14          & 50.10          & 44.15          & \textbf{55.82} & 44.33\\ \hline
T5-SQuAD$_1$ & Base  & 22.62          & \textbf{54.87} & \textbf{32.20} & \textbf{50.99} & \textbf{50.98} & \textbf{48.98} & \textbf{55.82} & \textbf{45.21}  \\ \hline
\end{tabular}
\label{tab:auto-evaluation}
\end{table}

Table \ref{tab:auto-evaluation} shows automatic evaluation comparison results with 
%ProphetNet \cite{qi2020prophetnet}, BART-HL and BART-SQuAD \cite{lewis-etal-2020-bart}; T5-HL and T5 \cite{raffel2020exploring}; and MixQG$_1$, MixQG$_2$, and MixQG-SQuAD \cite{murakhovska2021mixqg}. All but ProphetNet are pretrained on the base dataset and then finetuned on the SQuAD dataset. 
%\hl{ Changes start here: }
ProphetNet \cite{qi2020prophetnet}, BART \cite{lewis-etal-2020-bart}, T5 \cite{raffel2020exploring} and MixQG \cite{murakhovska2021mixqg}. 
BART-SQuAD, T5-SQuAD, and MixQG-SQuAD are corresponding models finetuned on the SQuAD dataset.
BART-hl and T5-hl are augmented models using the ``highlight" encoding scheme introduced by Chan and Fan \cite{chan2019recurrent}.
%\hl{ Changes end here}

The results of MixQG$_1$ were presented in the original paper \cite{murakhovska2021mixqg}, and the results of MixQG$_2$ were computed by us using the pretrained model posted on HuggingFace (https://huggingface.co/Salesforce/mixqg-base). %The result of MetaQG was our improved model.
The results show that, except BLEU, T5-SQuAD$_1$ outperforms all other models on the ROUGE and METEOR metrics,  produces the same BERTScore score as that of MixQG-SQuAD. 
%T5-SQuAD$_1$'s BLEU score is slightly lower than MixQG because some of the lengths of the generated questions are shorter which results in Brevity Penalty in the BLEU algorithm. 
 Overall, T5-SQuAD$_1$ performs better than all the models in comparison.

\subsection{Manual evaluations of TP3}

A number of publications (e.g., see \cite{callison-burch-etal-2006-evaluating,liu-etal-2016-evaluate,nema-khapra-2018-towards}) have shown that the aforementioned automatic evaluation metrics based on n-gram similarities do not always 
correlate well with human judgments about the answerability of a question. 
Thus, we'd also need to use human experts to evaluate the qualities of QAPs generated
by TP3. We do so 
%We then manually evaluated our base and large models 
on the Gaokao-EN dataset consisting of 85 articles, where each article contains 15 to 20 sentences. We chose Gaokao-EN because expert evaluations are provided to us from a project we work on.
%TP3 generates a total of at least 1,270 QAPs. 
Table \ref{tab:manually-evaluation} depicts the evaluation results. Title abbreviations are explained below,
where the numbers in boldface are the best in the corresponding columns:
\begin{enumerate}
\item \textbf{Total} means the total number of QAPs generated by TP3. 
\item \textbf{ADQT} means the total number of adequate QAPs. These QAPs can be directly used without any modification. 
\item \textbf{ACPT} means the total number of
QAPs where the question, while semantically correct, contains a minor English issue that can be corrected with a minor effort. For example,
a question may simply be missing a word or a phrase at the end. Such QAPs may be deemed acceptable. 
\item \textbf{UA} means unacceptable QAPs. 
\item \textbf{ADQT-R} means the ratio of the adequate QAPs over all the QAPs being generated.
\item \textbf{ACPT-R} means the ratio of the adequate and acceptable QAPs over all the QAPs being generated.
%\item \textbf{Better} means the QAPs among the adequate QAPs under the current model that are better than the adequate QAPs generated under different models with respect to the same answers.
\end{enumerate}

\begin{table}[h]
\centering
\caption{Manual evaluation results for TP3-Base and TP3-Large over Gaokao-EN}
\begin{tabular}{l|c|c|c|c|c|c|c}
\hline
\textbf{TP3} &	\textbf{Learning Rate} &	\textbf{Total} & \textbf{ADQT} & \textbf{ACPT}	 & \textbf{UA} 
	& \textbf{ADQT-R}	& \textbf{ACPT-R} %& \textbf{Better} 	
	\\ \hline
Base		           & 3e-5 & \textbf{1290}& 1145& \textbf{63}& 82& 88.76& 93.64 %& 20
\\ \hline
\multirow{5}{*}{Large} & 3e-5 & 1287& 1162& 49& 76& 90.29& 94.09 %& 19
\\\cline{2-8}
   					   & 1e-5 & 1271& 1166& 39& 76& 91.74& 94.81 %& 16
   					   \\\cline{2-8}
   					   & 8e-6 & 1270& 1162& 39& 69& 91.50& 94.57 %& 27
   					   \\\cline{2-8}
   					   & 6e-6 & 1273& \textbf{1172}& 45& \textbf{56}& \textbf{92.07}& \textbf{95.60} %& \textbf{36}
   					   \\\cline{2-8}
   				 & Dynamic & 1288& 1116& 51& 121& 86.65& 90.61 %& 22
   				 \\\hline
\end{tabular}
\label{tab:manually-evaluation}
\end{table} %Deep-learning-based Question Generation
\chapter{Sequence-learning-based Question Generation}

We present a learning scheme to generate QAPs via meta sequence representations of sentences. A meta sequence is a sequence of vectors comprising semantic and syntactic tags.

In particular, we use a sequence of vectors to represent a sentence, where each vector consists of a semantic-role (SR) tag, a part-of-speech (POS) tag, and other syntactic and semantic tags, and we refer to such a sequence as a \textsl{meta sequence}.
We then present a scheme called MetaQA to learn meta sequences of declarative sentences and the corresponding interrogative sentences from a training dataset. % consisting of such sentences.
Combining and removing redundant meta sequences yields a set called MSDIP (Meta-Sequence-Declarative-Interrogative Pairs), with each element being a pair of an MD and corresponding MI(s), where MD and MI stand for, respectively, a meta sequence for a declarative sentence and for an interrogative sentence.
A trained MetaQA model generates QAPs for a given declarative sentence $s$ as follows:
Generate a meta sequence for $s$, find a best-matched MD from MSDIP, generates meta sequences for interrogative sentences according to the corresponding MIs and the meta sequence of $s$, identifies the meta-sequence answer to each MI, and coverts them back to text to form a QAP.

Our objective is to generate adequate QAPs on a given declarative sentence written in a given language $L$. We %present a general framework 
%by assuming 
assume that $L$ has an oracle $O_L$ to
provide syntactic and semantic information on a given sentence.

\begin{enumerate}
\vspace*{-5pt}
\item  $O_L$ can distinguish simple sentences (i.e., there is only one predicate) and complex sentences (i.e., there are two or more predicates). A complex sentence has two kinds: The first kind consists of a simple sentence as a main clause and a few 
subordinate clauses (simple or complex sentences) or sentence segments
with  normalized verbs.
The second kind consists of 
a few independent sentences (simple or complex) connected by conjunction.

\vspace*{-5pt}
\item $O_L$ can segment sentences into a sequence of basic units.
A basic unit could be a phrasal verb, a phrasal noun, or simply a word 
that does not belong to any phrase (if any) contained in the sentence.

\vspace*{-5pt}
\item $O_L$ can assign each basic unit in a sentence with an SR tag and a POS tag.
For a complex sentence of the first kind, $O_L$ can tag the main clause as a simple sentence and each subordinate clause with one SR tag (such as time and cause), and tag each subordinate clause itself
as a sentence. For a complex sentence of the second kind, $O_L$ simply separates the sentence into a collection of individual sentences and tags them accordingly.
Moreover, $O_L$ may be able to produce  other semantic or syntactic tags for each basic unit. 

\vspace*{-5pt}
\item $O_L$ can identify an interrogative pronoun by a POS tag.
An interrogative sentence, however, may or may not include an interrogative pronoun. 
\end{enumerate}

For example, exiting NLP tools for
the English language provide a reasonable approximation to such an oracle. Better approximations are expected when more NLP techniques are developed.
%We suspect that any natural language would
%satisfy these assumptions as well. 
\begin{definition}
Let $k \geq 2$ be a number of tags that $O_L$ can assign to a basic unit.
A $k$-semantic-syntactic unit ($k$-SSU) is a $k$-dimensional vector of tags,
denoted by $(t_1, t_2, \ldots, t_k)$, where $t_1$ is an SR tag, $t_2$ is a POS tag, and $t_i$ ($i>2$) represent other tags of fixed types.
\end{definition}

For example, 
we may add an NE tag to a basic unit to form a 3-SSU; adding one more tag on sentiment  forms a 4-SSU. 
Let $U = (t_1,t_2,\ldots, t_k)$ be an SSU. Denote by $U.i = t_i ~(i \geq 1)$.
The prefix $k$ is omitted when there is no confusion. 

Two consecutive SSUs $A$ and $B$ with $A.1 = B.1$ (i.e., they have the same SR tag)
and $A$ appearing on the left side of $B$ in a sentence may be merged to a new SSU $C$ as follows:
(1) If $A = B$, then set $C \leftarrow A$.
(2) Otherwise, based on the underlying language $L$, either set $C.2 \leftarrow A.2$ (i.e., use the POS tag on the left) or set $C.2 \leftarrow B.2$.
For the rest of the tags in $C$, select a corresponding tag in $A$ or $B$ according to $L$. The following proposition is evident:

\begin{proposition} \label{prop:1}
For any sequence of SSUs, after merging, the new sequence of SSUs does not
have two consecutive SSUs with the same SR tag.
\end{proposition}

To accommodate the situation without proper segmentation of phrasal verbs,
% (see Section \ref{sec:4.7}), 
it is desirable to allow a fixed number of consecutive SSUs to have the same SR tag
in a meta sequence. 
 
\begin{definition} \label{def:2}
A {\itshape meta sequence} is a sequence  of SSUs such that 
each SR tag appears at most $r$ times,
with interrogative pronouns (if any) left as is without tagging, where
$r \geq 1$ is a positive constant. % depending on the underlying language.
\end{definition}

We assume the availability of \textsl{sentence segmentation} that can segment
%will only consider 
a complex sentence to form simple sentences for each clause (main and subordinate),
%with
%each clause being a simple sentence, 
and we treat such a sentence as 
a set of simple sentences. If a clause itself is a complex sentence, it can be further
segmented as a set of simple sentences. A declarative sentence consists of at least three different SR tags corresponding to
subject, object, and predicate.

Since a complex sentence can be treated as a list of simple sentences, 
MetaQA learns meta sequences 
of declarative sentences and the corresponding interrogative sentences from a training dataset
consisting of such pairs of sentences, where a declarative sentence is a simple
sentence. %After training, Meta generates QAPs on a given declarative sentence.

However, there are complex sentence that are not easily segmented into
a set of simple sentences using the existing NLP tools. To represent this type of complex sentences, we may
define a meta sequence as a recursive list of SSUs with a tree structure to represent a sentence using the notion of \textsl{list} in the LISP programming language. This
will be addressed in a separate paper.
% it is not needed for the purpose of this paper.

\begin{figure}[h]
  \centering
  \includegraphics[width=0.9 \linewidth]{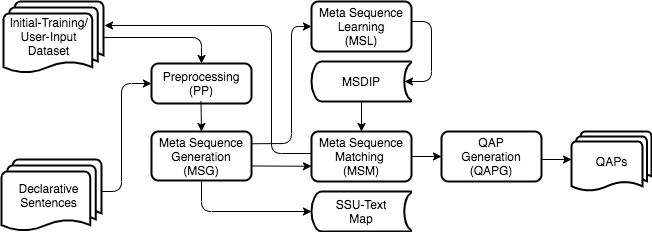}
  \caption{MetaQA architecture and data flow}
  \label{fig:3}
%   \Description{}
\end{figure}

MetaQA consists of two phases: learning and generation. In the learning phase,
MetaQA learns meta sequence pairs from an initial training dataset to generate an initial MSDIP.
In the generation phase, it takes a declarative sentence as input and 
generates QAPs using MSDIP.
Figure \ref{fig:3} depicts the general architecture and data flow of MetaQA,
which consists of six components: Preprocessing (PP), Meta Sequence Generation (MSG), Meta Sequence Learning (MSL), 
Meta Sequence Matching (MSM), and QAP Generation (QAPG) 
%(see Section \ref{sec:4} for detailed explanations of these components in connection to an implementation of the English language). 

Both phases use the same PP and MSG components. The PP component is responsible for
tagging basic units in a given sentence (declarative or interrogative) with SR tags, POS tags, and other syntactic and semantic tags, and segmenting complex sentences into a set of simple sentences using oracle $O_L$. The MSG component is responsible for merging SSUs to form a meta sequence. Moreover, for an input sentence in the generation phase, MSG also maps each SSU after merging to the underlying text. 

\section{Learning phase}   

%MetaQA
% first uses $O_L$ to assign SR tags, POS tags, and other syntactic and semantic tags
%to basic units in a given sentence (declarative or interrogative).
%It then merges consecutive SSUs 
%to form a meta sequence.
%Recall that an interrogative pronoun identified by POS tag is left as is without using its SSU.
The MSL component removes redundant meta sequences for each pair of
MD and MI generated from MSG and stores the remaining pairs in the MSDIP database.  
Recall that an interrogative pronoun identified by POS tag in an MI is left as is without using its SSU

%The learned result is a dataset called MSDIP (Meta-sequence
%Declarative-Interrogative Pairs). Inside each pair in MSDIP, the first element is
%a meta sequence for a declarative sentence (MD) and the second element is a set of
%meta sequences for the corresponding interrogative sentences (MI) in the dataset.

Note that for any language, $k$ is a constant, 
so are the number of SR tags, the number of POS tags, and
the number of other tags. 
The following proposition is straightforward.

\begin{proposition} \label{prop:2}
(1) For a given language, the length of a meta sequence is bounded above
by a constant, so is the size of MSDIP. 
(2) The length of a meta sequence for a declarative sentence
is at least 3. 
% is bounded above by a constant, so is 
\end{proposition}

\section{Generation phase} \label{sec:3}

Let $M$ be a meta sequence. Denote by
$M'$ the set of SSUs contained in $M$ and $|M|$ the size of $M'$.
After MetaQA is trained, it generates QAPs from
a given declarative sentences $s$ using the following
{\itshape QAP-generation algorithm}, where
$X_s$ is
the meta sequence for $s$ generated from MSG. Recall that the text
for each SSU is stored in the SSU-Text Map. 
%Note that $|X_s| \geq 3$ because $X_s$ must contain at least three SSUs with the corresponding SR tags being subject, object, and predicate.

Step 1. Find a meta sequence MD $X$ from (MD, MI) pairs in MSDIP 
that is the \textsl{best match} of $X_s$. 
This means that the longest common substring of $X$ and $X_s$,
denoted by $Z = \text{LCS}(X,X_s)$, is the longest 
among all MDs in MSDIP.  A substring is a sub-sequence of consecutive SSUs.
If $Z$ contains SSUs for, respectively, a subject, a predicate, and an object, then
we say that it is a \textsl{successful} match.
If furthermore, $Z = X = X_s$, then we say that it is  
a \textsl{perfect} match. If $Z$ is missing a subject SSU, a predicate SSU, or
an object SSU, then it is an \textsl{unsuccessful} matching.
If a match is successful, got Step 2.
If a match is unsuccessful or successful but not perfect, then notify the user that 
MetaQA needs to learn a new pattern and
ask for interrogative sentences for $s$ from the user. After this, go to Step 2.

% an unsuccessful match and a non-perfect successful match,

Step 2.
The goal is to generate all possible interrogative sentences for $s$.  
For each pair $(X,Y) \in \text{MSDIP}$,
generate a meta sequence $Y_s$ from $Y$ with
$$Y'_s = [Y'-(X' \cap Y'-X'_s)] \cup (X'_s - Z').$$
This means that $Y'_s$ is obtained from $Y'$ by removing SSUs that are in both
meta sequences in the matched pair
but not in the input sentence,
and adding SSUs in the input sentence but not in the matched MD.
Since $Z = \text{LCS}(X,X_s)$, the following proposition is straightforward:

\begin{proposition} \label{prop:3}
$X'_s - Z' = X'_s - X'$.
\end{proposition}
Order SSUs in $Y'_s$ appropriately to form $Y_s$, which requires
localization according to the underlying language.
If an SSU in $Y'_s$ has the corresponding text stored in Step 1,
then replace it. If not, then it requires localization to resolve it.
This generate an interrogative sentence $Q_{s}$ for $s$.

Step 3. For each interrogative sentence $Q_{s}$ generated in Step 3,
the SSUs in $A'_s = X' - Y'$ represent a correct answer.
Place SSUs in $A'_s$ in the same order as in $X'_s$ and replace each SSU with the corresponding text in $s$ to obtain an answer $A_s$ for $Q_{s}$.

% \begin{comment}
\section{An Implementation of MetaQA for English} \label{sec:4}

SR, POS, and NE tags are used in this implementation. Existing
NLP tools for generating these tags are for words, not for phrases. 
We could, however, use phrase segmentation to resolve this by appropriate merging operations.
While word segmentation is not needed %is a must in processing logographic languages, 
%there are ways to get around it 
in alphabetic languages such as English, 
phrase segmentation provides a better interpretation of the underlying sentence.
We first assume the existence of an ideal phrase segmentation for English,  
and then discuss how to get around it at the end of this section.

\subsection{Preliminaries}
The following NLP tools are used to generate tags: 
Semantic-Role Labeling (SRL) \cite{shi2019simple} for SR tags,
POS Tagging \cite{toutanova2003feature} for POS tags, and
Named-Entity Recognition (NER) \cite{peters2017semi}
for NE tags.

SR tags are defined in PropBank\footnote{https://verbs.colorado.edu/~mpalmer/projects/ace/EPB-annotation-guidelines.pdf} \cite{bonial2012english,martha2005proposition}, 
which consist of three types: ArgN (arguments of predicates), ArgM (modifiers or adjuncts of the predicates) , and V (predicates).
ArgN consists of six tags: ARG0, ARG1, $\ldots,$ ARG5, and
ArgM consist of
multiple subtypes such as
LOC as location, EXT as extent, DIS as discourse connectives, 
ADV as general purpose, NEG as negation, MOD as modal verb, CAU as cause, TMP as time, PRP as purpose, MNR as manner, GOL as goal, and DIR as direction.

POS tags$\,$\footnote{https://www.ling.upenn.edu/courses/Fall\_2003/ling001/penn\_treebank\_pos.html} are defined in the Penn Treebank tagset \cite{toutanova2003feature,marcus1993building}. For example, NNP is for singular proper noun,
VBZ for third-person-singular-present-tense verb,
DT for determiner, and IN for preposition or subordinating conjunction.

NE tags include PER for persons, ORG for organization, LOC for locations, and numeric expressions for time, date, money, and percentage.

\subsection{PP, MSG, and MSL Localization}

The PP, MSG, and MSL components, on top of what is described in
Section \ref{sec:3}, incur the following localization.
%
% are the same as in
%Section 3, except the following localization in preprocessing:
%
%\subsection{Preprocessing}
%
%The PP component first 
PP first replaces contractions and slang with
words or phrases to help improve tagging accuracy.
For example,  contractions \textsl{'m, 's, 're, 've, n't, e.g., i.e., a.k.a.} are replaced by,
respectively, \textsl{am, is, are, have, not, for example, that is, also known as}.
Slang \textsl{gonna, wanna, gotta, gimme, lemme, ya}  are replaced by,
respectively, \textsl{going to, want to, got to, give me, let me, you}.

PP then segments sentences and
tags words in sentences using SRL, POS Tagging, and NER
for the training dataset and later for input sentences for generating QAPs.
Use SRL to segment a complex sentence into a set of simple sentences and discard all simple sentences without 
a subject or an object. Note that there are complex sentences that are hard to segment using SRL.
Moreover, for each sentence, PP removes
all the words with a CC (coordinating conjunction) as POS tag before its subject, including
\textsl{and, but, for, or, plus, so, therefore}, and \textsl{because}.

%\subsection{Meta Sequence Generation}

%The MSG component generates a meta sequence for a sentence
%in the learning phase and a meta sequence for a declarative sentence in the 
%generation phase.

%Every word in a sentence has an SR tag and a POS tag from preprocessing,
%but NE tags may be null. 
% Then MSG merges the SSUs for words in each basic unit as follows:
% (1) If the unit contains a V-SSU (i.e., an SSU with V as the SR tag),
% then use this V-SSU for the entire unit.
% (2) If the unit contains no V-SSU, then it must contain a noun,
% use the first SSU from the right 
% with a noun POS tag.

MSG then merges the remaining SSUs if two consecutive SSUs are identical.
If they are not identical but have the same SR tag,
then use this SR tag in the merged SSU, and
the POS tag in the first SSU from the right. 
If they contain a noun, use the first SSU from the right with a noun POS tag.
Moreover, the NE tag in the merged SSU is null
if both SSUs contain a null NE tag; otherwise,
use the first non-empty NE tag from the right. 

%For an input sentence in the
%generation phase,  MSG also stores a mapping between a merged SSU and the
%corresponding text in the SSU-Text Map.

%\subsection{Meta Sequence Learning} 
%The MSL component removes redundant meta-sequence pairs generated for the training dataset, with each pair in the form of (MD, MI), where MD representing the meta sequence of
%a declarative sentence and MI the meta sequence for the corresponding interrogative sentence. %Not that there may be multiple MIs for the same MD. The remaining meta-sequence
%pairs are stored in the MSDIP file, which is typically small, and bounded above by
%a constant (Proposition \ref{prop:2}).

\subsection{MSM Localization}
The MSM component takes a meta sequence $X_s$ of a sentence $s$ as input 
and executes Step 1 in the
QAP-generation algorithm described in Section \ref{sec:3} using Ukkone's Suffix-Tree algorithm \cite{ukkonen1985algorithms} to compute a longest common substring of two
meta sequences, which runs in linear time. During matching, the POS tags for various types of nouns 
are treated equal; they are NN, NNP, NNS, and NNPS, The POS tags 
for third-person-singular-present verbs are treated equal; they are VBP and VBZ.
To use Ukkone's algorithm,,
we encode a meta sequence as a sequence of symbols using $/$ to separate tags in an SSU.
That is, vector $(t_1,t_2,t_3)$ is now written as $t_1/t_2/t_3$. If $t_2$ is null, then write it as
$t_1//t_3$. If $t_3$ is null, then write it as $t_1/t_2/$. If both are null, then write it
as $t_1//$. SSUs in a sequence are just written as concatenation. For example,
the sentence ``Abraham Lincoln the 16th president of the United States" has
the following SSUs:

\textrm{Abraham (ARG1/NNP/PER) Lincoln (ARG1/NNP/PER) was (V/VBZ/) the (ARG2/DT/)
16th (ARG2/JJ/) president (ARG2/NN/) of (ARG2/IN/) the (ARG2/DT/) United (ARG2/NNP/LOC)
States (ARG2/NNP/LOC)}.

The meta sequence for this sentence is, after merging: 
\textrm{ARG1/NNP/PER V/VBZ/ ARG2/NNP/LOC}.

Let $X$ be an MD in MSDIP such that LCS($X,X_s$) is the longest
among all MDs in MDDIP,
denoted by $Z$. 

%If matching is perfect, move to the QAPG component.
%If matching is not successful, then
%move to the UEIS component,
%inform the user that a new pattern is encountered, and ask for input of
%interrogative sentences to train new pairs for MSDIP.
%If matching is successful but not perfect, then the sentence $s$ still represents a new pattern %not yet learned, even though a question may still be generated.
%In this case, MSM moves to QAPG, as well as to UEIS as an option for new training.

\subsection{QAPG Localization}

The QAPG component executes Steps 2--3 in the QAP-generation algorithm described in
Section \ref{sec:3}. Recall that $Z = \text{LCS}(X,X_s)$ is the longest match
among all MDs in MSDIP, and
after 
the set of SSUs $Y'_s$ is generated, localization is needed to form $Y_s$.

\textsl{Case 1:} $Z=X_s$. Then $Y_s = Y$. 
%Replace each SSU in $Y_s$ with the corresponding text stored in Step 1. 

\textsl{Case 2:} $Z$ is a proper substring of $X_s$. Then each SSU in $X'_s - Z$ appears either 
before $Z$ or after $Z$. Form a string $Y_b$ and $Y_a$ of the SSUs that appear,
respectively, before and after $Z$ in the same order as they appear in $X_s$.
Let $Y_s =  [Y-(X'\cap Y'-X'_s)]Y_aY_b$, where $Y-(X'\cap Y'-X'_s)$
means to remove from $Y$ the SSUs in $X'\cap Y'-X'_s$.
 
For each SSU in $Y_s$ if a corresponding text can be found in the SSU-Text Map,
then replace it with the text. An SSU that doesn't have a matched text in the
SSU-Text Map is
due to the helping verbs added in the interrogative sentence that generates $Y$.
%
%Some SSUs in $Y_s$, however, may not have a match in $X_s$ and so
%cannot be replaced with appropriate text directly from the SSU-Text Map.
%Since each declarative sentence must include a subject, an object, and a predicate, there %must be texts for the SSUs that represent these entities in the SSU-Text Map. 
There are five POS tags for verbs: VBG for gerund or present participle,
VBD past tense, VBN past participle, VBP non-3rd person singular present,
and VBZ 3rd person singular present.
Present participle and past participle have already included helping verbs, and so do the negative forms of
past tense and present tense.
Thus, only positive forms of past tense (VBD) and present tense (VBP, VBZ) 
do not include
helping verbs, which need to be resolved.

\section*{Rule for resolving helping verbs}

The first V-SSU in $Y$ (i.e., the SSU that contains the SR tag of V) is a helping verb. To determine its form, 
check the POS tag in the subject SSU (usually it is ARG0)
% check the POS tag in the ARG0-SSU 
and determine if it is singular or plural. Then check the POS tag in the first V-SSU in $Y$ 
to determine the tense. Replace the second V-SSU with the verb in its original form 
for the V-SSU in the SSU-Text MAP. 

For example, suppose that the following declarative sentence ``John traveled to Boston last week" and its
interrogative sentence about location ``Where did John travel to last week"
are in the training dataset, which generate the following SSUs before merging:

John (ARG0/NNP/PER) traveled (V/VBD/) to (ARG1/IN/) Boston (ARG1/NNP/LOC) last (TMP/NN/) week (TMP/NN/).

Where (Where) did (V/VBD) John (ARG0/NNP/PER) travel (V/VB/) to (ARG1/IN/) 
last (TMP/NN/) week (TMP/NN/)?

Since ``travel to" is a phrasal verb, after merging, we have

John (ARG0/NNP/PER) traveled  to (V/VBD/) Boston (ARG1/NNP/LOC) last week (TMP/NN/).

Where (Where) did (V/VBD) John (ARG0/NNP/PER) travel to (V/VB/) last week (TMP/NN/)?

The following meta-sequence pair $(X,Y)$ is learned for MSDIP:

X  = \textrm{ARG0/NNP/PER V/VBD/ ARG1/NNP/LOC TMP/NN/}

Y = \textrm{Where V/VBD/ ARG0/NNP/PER V/VB/ TMP/NN/}

Suppose that we are given a sentence $s=$ ``Mary flew to London last month." Its meta sequence $X_s$ is exactly the same as $X$, with
ARG0/NNP/PER for ``Mary",
V/VBD/ for ``flew to", ARG1/NNP/LOC for ``London", and TMP/NN/ for ``last month", which
are stored in the SSU-Text Map.
Thus, $Y_s = Y$. 
We can see that 
the SSU of V/VB/ %and ARG1/IN/ 
in $Y$ is not in the SSU-Text Map.
To resolve the unmatched V/VB/, check the POS tag in the ARG0-SSU, which is NNP, indicating a singular noun.
The POS tag in the first V-SSU is VBD, indicating past tense. Thus, the correct form of the helping verb is ``did". The text for V/VBD is ``flew to" in the SSU-Text Map. The original form of the verb is ``fly".
Thus, the second V-SSU is replaced with ``fly". 
%The remaining text ``to" is for ARG1/IN/.
This generates the following interrogative sentence: ``Where did Mary fly to last month?"
The answer SSU is $X'-Y'$, which is ARG1/NNP/LOC, corresponding to ``London".

\subsection{SSU Merging without Segmentation} \label{sec:4.7}

To the best of our knowledge, no tools exist at this point that can segment English sentences to identify phrasal nouns and phrasal verbs. It is worth mentioning that AutoPhrase \cite{shang2018automated} 
can be used for identifying certain phrasal nouns. 
We could deal with 
phrasal verbs using a list of common phrasal verbs or by modifying merging operations.
A phrasal verb consists of a preposition or an adverb, or both. 
There are four POS tags  IN for preposition or subordinating conjunction,
RB for adverb, RBR for comparative adverb, and RBS for superlative adverb.

To see this problem, let us look at the same example aforementioned. After merging,
we have

John (ARG0/NNP/PER) traveled (V/VBD/) to Boston (ARG1/NNP/LOC) last week (TMP/NN/).

Where (Where) did (V/VBD/) John (ARG0/NNP/PER) travel (V/VB/) to (ARG1/IN/) last week (TMP/NN/)?

For the input sentence we have

Mary (ARG0/NNP/PER) flew (V/VBD/) to London (ARG1/NNP/LOC) last moth (TMP/NN/).

The interrogative sentence is ``Where did Mary fly ARG1/IN/ last week?" after replacing SSUs with text in
the SSU-Text Map, with ARG1/IN/ unmatched with text.
We can resolve this by modifying the merging operation as follows:
When an SSU with a POS tag for preposition or adverb appears appears before or after a V-SSU, leave it as is without merging it with its neighboring SSUs of the same SR tag, unless the POS tags 
in them are also for prepositions or adverbs. The rest of the merging operations are the same. Then we have, after merging,

John (ARG0/NNP/PER) traveled (V/VBD/) to (ARG1/IN/) Boston (ARG1/NNP/LOC) last week (TMP/NN/).

Where (Where) did (V/VBD/) John (ARG0/NNP/PER) travel (V/VB/) to (ARG1/IN/) last week (TMP/NN/)?

Now the input sentence becomes, after SSU merging,

Mary (ARG0/NNP/PWR) flew (V/VBD) to (ARG1/IN/) London (ARG1/NNP/LOC) last moth (TMP/NN/).

All the SSUs in the meta sequence  ``Where V/VBD/ ARG0/NNP/PER V/VB/ ARG1/IN/ TMP/NN/" have corresponding text 
 in the SSU-Text Map after resolving for helping verbs.  The answer
SSU is in $X'-Y'=$ ARG1/NNP/LOC, which is ``London". 
% \end{comment}

% \begin{comment}
\section{Running Samples}  \label{sec:examples}

\paragraph{Example 1.}
Suppose that the following declarative sentence and the corresponding
two interrogative sentences are given as training data at the learning phase:

``Amanda has a story book on the American history." 
 
``Who has a story book on the American history?" 

``What does Amanda have?"

PP generates corresponding SSUs as follows:

\textrm{
Amanda (ARG0/NNP/PER) has (V/VBZ/) a (ARG1/NN/) nice (ARG1/NN/) textbook (ARG1/NN/) on (ARG1/NN/) programming (ARG1/NN/).}

\textrm{Who (Who) has (V/VBZ/) a (ARG1/NN/) nice (ARG1/NN/) textbook (ARG1/NN/) on (ARG1/NN/) programming (ARG1/NN/)?}

\textrm{What (What) does (V/VBZ/) Amanda (ARG0/NNP/PER) have (V/VB/)?}

MSG merges SSUs and MSL places two pairs $(X_1, Y_{1,1})$ and $(X_1, Y_{1,2})$ in MSDIP if they are
not already present, where

$X_1 =$ \textrm{ARG0/NNP/PER V/VBZ/ ARG1/NN/, }

$Y_{1,1} =$ \textrm{Who V/VBZ/ ARG1/NN/}, 

$Y_{1,2} =$ \textrm{What V/VBZ/ ARG}. 

Suppose that the following two declarative sentences are given at the generation phrase:

$s_1=$ ``Tom has a story book on the American history."

$s_2=$ ``Duncan Watts agrees with the conclusion."

PP generates SSUs as follows, with ``agrees with" recognized as a phrasal verb.

\textrm{Tom (ARG0/NNP/PER) has (V/VBZ/) a (ARG1/NN/) story (ARG1/NN/) book (ARG1/NN/) on (ARG1/NN/) the (ARG1/NN/) American (ARG1/NN/) history (ARG1/NN/)}.

\textrm{Duncan (ARG0/NNP/PER) Watts (ARG0/NNP/PER) agrees (V/VBZ/) with (ARG1/IN/) the (ARG1/NN/) conclusion (ARG1/NN/)}.

MSG merges SSUs, generates the following two meta sequences:

$X_{s_1} = X_{s_2}=$ \textrm{ARG0/NNP/PER V/VBZ/ ARG1/NN/}, 

and
places the following in the
SSU-Text Map:

\textrm{Tom (ARG0/NNP/PER) has (V/VBZ/) a story book on the American history (ARG1/NN/)}.

\textrm{Duncan Watts (ARG0/NNP/PER) agrees with (V/VBZ/) the  conclusion (ARG1/NN/)}.

MSM  finds a match
$X_1 = X_{s_1} = X_{s_2}$. QAPG
generates

$Y_{s_1,1} = Y_{s_2,1}=$ \textrm{Who V/VBZ/ ARG1/NN/},

$Y_{s_1,2}= Y_{s_2,2}=$ \textrm{What V/VBZ/ ARG0/NNP/PER V/VB/}.

The corresponding interrogative sentences are, after applying the rule for resolving VB for
$Q_{s_1,2}$ and $Q_{s_2,2}$:

$Q_{s_1,1} =$ ``Who has a story book on the American history?"

$Q_{s_1,2}=$ ``What does Tom have?"

$Q_{s_1,1} =$ ``Who agrees with the conclusion?"

$Q_{s_1,2}=$ ``What does Duncan Watts agree with?"

The SSU for the answer to $Y_{s_1,1}$ is ARG0/NNP/PER with ``Tom" being
the text and 
the answer to $Y_{s_1,2}$ is ARG1/NN/ with ``a story book on American history" being the 
text. Likewise, the SSU for the answer to $Y_{s_2,1}$ is ARG0/NNP/PER with
``Duncan Watts" being the text and the answer to $Y_{s_2,2}$ is ARG1/NN/ with
``the conclusion" being the text.

\textsl{Remark.} Using the modified merging operation without using segmentation, we have$X_{s_2}=$ ARG0/NNP/PER V/VBZ/ ARG1/IN/ ARG1/NN/, and so 
$X$ is no longer a successful match. A new pattern is needed to learn for MSDIP.

% \begin{comment}
\paragraph{Example 2.}
Suppose that the following two pairs of declarative and interrogative sentences
are in the training dataset: 

``A doughnut is a fried dough confection."

``What is a doughnut?"

``Uranus is a unusual planet because it is tilted."

``Why is Uranus an unusual planet?"

The SSUs for these sentences are

\textrm{A (ARG1/NN/) doughnut (ARG1/NN/) is (V/VBZ/) a (ARG2/NN/) fried (ARG2/NN/) dough (ARG2/NN/) confection (ARG2/NN/).}

\textrm{What (What) is (V/VBZ/) a (ARG2/NN/) doughnut (ARG2/NN/)?}

\textrm{Uranus (ARG1/NNP/) is (V/VBZ/) an (ARG2/NN/) unusual (ARG2/NN/) planet (ARG2/NN/) because (CAU/VBN/) it (CAU/VBN/) is (CAU/VBN/) tilted (CAU/VBN/).} 

\textrm{Why (Why) is (V/VBZ/) Uranus (ARG1/NNP/) an (ARG2/NN/) unusual (ARG2/NN/) planet (ARG2/NN/)?}

After merging SSUs, the following two meta-sequence pairs are learned:

$(X_2 Y_2) =$ (\textrm{ARG1/NN/ V/VBZ/ ARG2/NN/}, \textrm{What V/VBZ/ ARG2/NN/}),

$(X_3, Y_3) =$ (\textrm{ARG1/NNP/ V/VBZ/ ARG2/NN/ CAU/VBN/}, \textrm{Why V/VBZ/ ARG1/NNP/ ARG2/NN/}).

Now suppose $s_3=$ ``The solar panel manufacturing industry is in the doldrums because supply far exceeds demand" is an input sentence for generating QAPs, which has
the following SSUs:

\textrm{The (ARG1/NN/) solar (ARG1/NN/) panel (ARG1/NN/) manufacturing (ARG1/NN/) industry (ARG1/NN/) is (V/VBZ/) in (ARG2/NN/) the (ARG2/NN/) doldrums (ARG2/NN/) because (CAU/NN/) supply (CAU/NN/) far (CAU/NN/) exceeds (CAU/NN/) demand (CAU/NN/).}

After merging SSUs we have a meta sequence $X_{s_3} =$ \textrm{ARG1/NN/ V/VBZ/ ARG2/NN/ CAU/NN/} with the following SSU-Text Map:

\textrm{The solar panel manufacturing industry (ARG1/NN/) is (V/VBZ/) in the doldrums (ARG2/NN/) because supply far exceeds demand (CAU/NN/)}.

We can see that $\text{LCS}(X_2, X_{s_3}) = \text{LCS}(X_3, X_{s_3}) =$ ARG1/NN/ V/VBZ/ ARG2/NN/, which generates the following two meta sequences:

$Y_{s_3,1} =$ (\textrm{What V/VBZ/ ARG2/NN/}),

$Y_{s_3,2} =$ (\textrm{Why V/VBZ/ ARG1/NNP/ ARG2/NN/ CAU/VBN/, where CAU/VBN/})
$\in X'_{s_3} - X'_2$ is added to $Y_3$ to get $Y_{s_3,2}$.

The corresponding interrogative sentences are, after applying the rule for resolving helping verbs:

$Q_{Y_{s_3,1}} =$ ``What is in the doldrums because supply far exceeds demand?"

$Q_{Y_{s_3,2}}=$ ``Why is the solar panel manufacturing industry in the doldrums?"

The answer to $Q_{Y_{s_3,1}}$ is the text for SSU $\in  X'_2-Y'_2 =$ ARG1/NN/, which is
``The solar panel manufacturing industry."
Likewise, the answer to $Q_{Y_{s_3,2}}$ is the text for SSU $ \in X'_3-Y'_3=$ CAU/NN/,
which is ``because supply far exceeds demand."
% \end{comment}

\section{Evaluations}

To evaluate MetaQA, we need to have appropriate evaluation measures,
training data, and evaluation data.
BLUE \cite{10.3115/1073083.1073135}, ROUGE \cite{lin-2004-rouge}, and Meteor \cite{10.1007/s10590-009-9059-4} are standard evaluation metrics for measuring automatic summarization and machine translation, which are good for computing text similarity
and have also been used to evaluate QG. Another commonly-used measure is 
human judgments.
BLEU and ROUGE-N count the number of overlapping units between the candidate text and the reference text by using N-grams.
ROUGE-L measures the cognateness between the candidate text and the reference text by using Longest common sub-sequence.
Meteor compares the candidate text with the reference text in terms of exact, stem, synonym, and paraphrase matches between words and phrases.
%It was shown that these metrics are either weak or have no correlation with human judgments  \cite{callison-burch-etal-2006-evaluating,liu-etal-2016-evaluate} . 
These metrics, however, do not evaluate grammatical correctness. Thus, human judgment 
is the only liable measure for grammatical correctness.

SQuAD \cite{Rajpurkar_2016} is a  dataset that has been
used for training and evaluating generative methods for QG.
However, not all QAPs in SQuAD are well-formed or with correct answers. 
There are also about 20\% of questions in the dataset that require paragraph-level information.
Thus, SQuAD is unsuitable for evaluating QAPs for our purpose.
Instead, we constructed an initial training dataset by writing a number of declarative sentences
and the corresponding interrogative sentences to cover the major
tense, participles, voice, modal verbs, and some common phrasal verbs such as
``be going to" and ``be about to" for the following six interrogative pronouns: 
\textsl{Where, Who, What, When, Why, How many}. A total of 112 meta-sequence pairs (MD, MI) were learned as  the initial MSDIP.

To evaluate MetaQA, we extracted declarative sentences from the official SAT practice reading tests$\,$\footnote{https://collegereadiness.collegeboard.org/sat/practice/full-length-practice-tests}, for the reason that
SAT practice reading tests provide a large number of different patterns of declarative sentences. 
 There are a total of eight SAT practice reading tests, each consisting of five articles and
 each article consisting of around 25 sentences, for a total of 40 articles and 1,136 sentences. 
After removing easy-to-identify interrogative sentences and imperative sentences,
we harvested a total of 1,025 sentences (which may still contain imperative
sentences).
Using the initial MSDIP, MetaQA generated a total of 796 QAPs.
% with the following breakdowns:

Three native Chinese speakers evaluated the QAPs on a shared
Google doc file based on the following criteria:
For questions: Check both syntax and semantics: (1) correct; (2) acceptable (e.g., a minor would make it correct); (3) not acceptable.
 For answers: (1) matched---the answer matches well with the question; (2) acceptable; (3) not acceptable.
 The final results were agreed by the three judges. Presented below 
 are questions generated with detailed breakdowns in each
category, where ``all correct" means both syntactically and semantically correct and conforming to native-speaker norms,
``not acceptable" means either syntactically or semantically unacceptable, and
``How" means ``How many": 

\medskip
\begin{center}
\begin{tabular}{l|c|c|c|c|c|c|c}
\hline
& \textbf{Where} & \textbf{Who} & \textbf{What} & \textbf{When} & \textbf{Why} & \textbf{How} & \textbf{Total} \\
\hline
MSDIP pairs & 18 & 45 & 23 & 22 & 6 & 8 & 122 \\
\hline
QAPs generated & 26 & 216 & 466 & 51 & 15 & 22 & 796 \\
\hline
All correct & 21 & 208 & 458 & 51 & 15 & 20 & 773 \\
\hline
Syntactically acceptable & 4 & 4 & 3 & 0 & 0 & 2 & 13 \\
\hline
Semantically acceptable & 1 & 2 & 5 & 0 & 0 & 0 & 8	 \\
\hline
Not acceptable & 0 & 2 & 0 & 0 & 0 & 0 & 2 \\
\hline
\end{tabular}
\end{center}
\smallskip
The percentage of generated questions that are both syntactically and semantically correct
is 97\%. %Among the 773 all-correct questions, their answers are also all correct. 
We noticed that there is a strong correlation between
the correctness of the questions and their answers. In particular, 
when a generated question is all correct, its answer is also all correct.
When a question is acceptable, its answer may be all correct or acceptable.
Only when a question is unacceptable, its answer is also unacceptable.
 
%The evaluation result is 97\%. 
%Mismatched QAs are due to inaccuracy of NER, 

The 13 incorrect but syntactically acceptable questions are mostly due to some minor issues
in segmenting a complex sentence into simple sentences, where a better
handling of sentence segmentation is expected to correct these issues. Two questions whose
interrogative pronoun should be ``how much" are mistakenly using ``how many".
Further refinement of POS tagging that distinguish uncountable nouns from countable nouns would solve this problem. The eight semantically acceptable questions are all due to 
NE tags that cannot distinguish between persons, location, and things. Further refinement of NE tagging will solve this problem. The two unacceptable questions are due to
serious errors induced when segmenting complex sentences. This suggests that we should
look into using a recursive list to represent complex sentences.

There were 589 sentences for which no matched meta sequences are found from 
the initial MSDIP. By learning new meta sequences from user inputs,
535 of these sentences found perfect matching, which generate
QAPs that are both syntactically and semantically correct. For the remaining 84 sentences,
some of then are imperative sentences without a clear structure of subject-predicate-object,
and some are hard to segment into a set of simple sentences due to inaccurate SR tagging
and so no
appropriate (MD, MI) pairs were learned. This suggests that we should
look into better sentence segmentation methods or meta trees as recursive lists of meta sequences to represent complex sentences
as a whole, which is left for future work.
 %Sequence-learning-based Question Generation
\chapter{Distractor Generation}

\begin{figure}[!h]
  \centering
  \includegraphics[width= \linewidth]{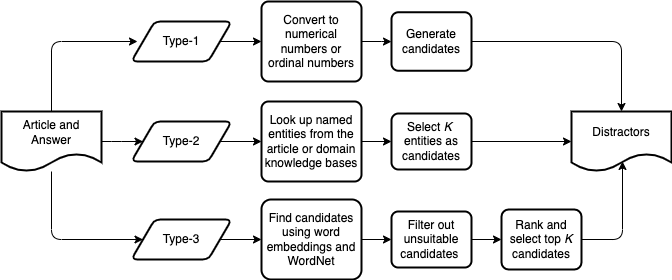}
  \caption{Distractor generation flowchart}
  \label{fig:dgf}
%   \Description{}
\end{figure}

Our distractor generation method takes the original article and the answer as input, and generates distractors as output. Figure~\ref{fig:dgf} depicts the data flow of our method.

Answers in QAPs are classified into two kinds. The first kind consists of just a single target word while the second kind consists of multiple target words. The latter is the case when  the
answer is a sentence or a sentence segment. 

For an answer of the first kind, if it is a type-1 or type-2 target, we use the methods
described in Section \ref{sec:distractors2} to generate three distractors; if it is a type-3 target, we use
the method described in Section \ref{sec:distractors3} to generate distractor candidates.
If there are at least three candidates, then select three candidates
with the highest ranking scores. 
%If there are less than three candidates, then we may fail to
%generate an adequate MCQ for this QAP.

For an answer of the second kind, for each type of a target word contained in it, we use the methods described in both Sections \ref{sec:distractors2} and
\ref{sec:distractors3} to generate distractors for target words in a fixed ordered preference of subjects, objects, adjectives for subjects, adjectives for objects,
predicates, adverbs, which can be obtained by semantic-role labeling.
Target words are replaced
according to the following preference:
%
%If there are more than three distractors, then select
%three distractors according to the following preference:
type-1 temporal, type-1 numerical value, type-2 person, type-2 location,
type-2 organization, type-3 noun (phrasal noun), type-3 adjective, type-3 verb (phrasal verb), and type-3 adverb. 

If the number of distractors for a given preference is less than three, then we generate extra distractors for a target word in the next preference. If all preference is gone through we still need more distractors, we could extend the selection threshold values to allow more candidates to be selected.

\section{Distractors for Type-1 and Type-2 Targets}
\label{sec:distractors2}

If a type-1 target is a point in time, a time range, a numerical number, an ordinal number, or anything that can be converted to a numerical number or an ordinal number (e.g. Friday may be converted to 5), which can be
recognized by regular expressions based on a POS tagger, then we devise several algorithms to alter time and number, and randomly select one of these algorithms when generating distractors. For example, we may  increase or decrease the answer value by one or two units, change the answer value at random from a small range of values around the answer,
or simply change the answer value at random. If a numerical value or an ordinal number
is converted from a word, then the result is converted back to the same form. For example,
suppose that the target word is ``Friday", which is converted to a number 5. If the distractor
is a number 4, then it is converted to Thursday. 

If a type-2 target is a person, then we first look for different person names that appear in the article using an NE tagger to identify them, and then randomly choose a name as a distractor. 
If there are no other names in the article, 
then we use Synonyms (http://www.synonyms.com) or a domain knowledge base on notable people we constructed to find a distractor. 
If a type-2 target is a location or an organization, we find a distractor in the same way by first
looking for other locations or organizations in the article, and then using
Synonyns and domain knowledge bases to look for them if they cannot be found in
the article.
For example, If the target word is a city, then a distractor should also  be city that is "closely" related to the target word. Distractors to the answer word "New York" should be cities in the same league, such as ``Boston", ``Philadelphia", and ``Chicago".  

\section{Distractors for Type-3 Targets}
\label{sec:distractors3}
\noindent
%For a given answer in a QAP, we may choose a (phrasal) noun, (phrasal) verb, adjective, or adverb contained in it in a fixed order (e.g., from right to left) as a sequence of target words. 
For a type-3 target, %which could be a (phrasal) noun or an adjective,
we find distractor candidates 
using word embeddings %antonyms,
%Word2vec \cite{10.5555/2999792.2999959} 
with similarity in a threshold interval (e.g.,[0.6,0.85]) so that
a candidate is not too close nor too different from the correct answer
and %then find synonyms and %Antonyms, and 
hypernyms using WordNet \cite{10.1145/219717.219748}. 
Note that a similarity interval of $[0.6,0.85]$ for word embeddings often
include antonyms of the target word, and we can use WordNet or an online dictionary to determine
antonyms.

Not all distractor candidates are suitable. Thus, we first filter out unsuitable candidates as follows:
\begin{enumerate}
\item Remove distractor candidates that contain the target word, for it may be
too close to the correct answer. For example, if 
``breaking news" is
a generated distractor candidate for the target word "news", then it is
removed from the candidate list since it contains the target word.
%the "Breaking news" need to be excluded.
% 1.	排除distractor  candidates中包含target word的candidates，例如 breaking news 中包含news，所以排除breaking news

\item Remove distractor candidates that have the same prefix of the target word with edit distance less than three, for such candidates may often be misspelled words from the target word.
For example, suppose the target word is "knowledge", then 
Word2vec may return a misspelled candidate "knowladge" with a high similarity,
which should be removed.
% 2.	排除前缀prefix相同且edit distance 小于3的单词
\end{enumerate}

We then rank each remaining candidate using the following measure:
\begin{enumerate}
\item Compute the Word2vec cosine similarity score $S_v$ for each distractor candidate $w_c$ with the 
target word $w_t$. Namely, $$S_v = \mbox{sim}(v(w_c),v(w_t)),$$
where $v(w)$ denotes a word embedding of $w$.
%If the candidate $w_c$ is an acronym of the target word, then let
%$$S_v = \frac{1}{\mbox{sim}(w_c,w_t)}.$$
%If no distractor candidates or an insufficient number of distractor candidates are found in a Word2vec dataset, then we use the WordNet Wu-Palmer (WUP) \cite{10.3115/981732.981751} score as the similarity score.
% 3.	计算每个 distractor candidates 和 target word 的word2vec similarity score Sv， 如果word2vec中查询不到这个candidate，则取 WUP score作为 similarity score

\item Compute the WordNet WUP score \cite{10.3115/981732.981751} $S_n$ for each distractor candidate with the target word. If the distractor candidate cannot be found in the WordNet dataset, set the WUP score to 0.1
for the following reason: If a word with a high score of word-embedding similarity to the target word but does not exist in the WordNet dataset, then it is highly likely that the word is misspelled, and so its ranking score should be reduced.
% 4.	计算每个distractor candidates 和 target word 的 WordNet Wu-Palmer (WUP) score Sn，如果WordNet中查询不到这个candidate，则取 WUP socre 为 0.1， 因为在word2vec中similarity score较高但在WordNet中不存在的单词，很大的可能性是拼写错误的单词，应该降低他的排序。例如knowledge和knowladge(e拼错成了a)，newspaper和newpaper。

\item Compute the edit distance score $S_d$ of each distractor candidate with target word by the
following formula:
\begin{equation*}\label{eq1}
    S_d = 1 - \frac{1}{1 + e^E},
\end{equation*}
where $E$ is the edit distance. Thus, a lager edit distance $E$ results in a smaller score $S_d$.

\item Compute the final ranking score $R$ for each distractor candidate $w_c$ with respect to the target word $w_t$ by
\begin{align*}\label{eq2}
R'(w_c,w_t) &= \left\{
\begin{array}{ll}
\frac{1}{4}(2S_v + S_n + S_d), &\mbox{if $w_c$ is an} \\
							&\mbox{antonym of $w_t$}, \\
\frac{1}{3}(S_v + S_n+S_d), & \mbox{otherwise},
\end{array}\right. \\
R(w_c,w_t) &=  - R'(w_c,w_t)\log R'(w_c,w_t).\\
\end{align*}
Note that $S_v, S_n, S_d$ are each between 0 and 1, and so
$R'(w_c,w_t)$ is between 0 and 1, which implies that $- \log R'(w_c,w_t) > 0$.
%Also note that $\log R'(w_c,w_t)$ results in a larger value of 
Also note that we give more weight to antonyms. 
% 6.	计算每个distractor candidates 和 target word 的Ranking score，
% R = Sv * Sn * Sd

\end{enumerate}

\section{Running Samples}
Given below are a few adequate MCQs with automatically generated distractors by our method:

\subsubsection*{Example 1}
Question: What does no man like to acknowledge? (SAT practice test 2 article 1)

Correct answer: that he has made a mistake in the choice of his profession.

Distractors:
\begin{enumerate}
\item that he has made a mistake in the choice of his association.
\item that he has made a mistake in the choice of his engineering.
\item that he has made a mistake in the way of his profession.
\end{enumerate}

\subsubsection*{Example 2}

When should ethics apply? (SAT practice test 2 article 2)

Correct answer: 
when someone makes an economic decision.

Distractors:
\begin{enumerate}
%\item when someone makes an economic consideration.
\item when someone makes an economic request.
\item when someone makes an economic proposition.
\item when someone makes a political decision.
\end{enumerate}

\subsubsection*{Example 3}

Question: 
What did Chie hear? (SAT practice test 1 article 1)

Correct answer: 
her soft scuttling footsteps, the creak of the door.

Distractors:
\begin{enumerate}
\item her soft scuttling footsteps, the creak of the driveway.
\item her soft scuttling footsteps, the creak of the stairwell.
\item her soft scuttling footsteps, the knock of the door.
\end{enumerate}

\subsubsection*{Example 4}
Question: 
Who might duplicate itself? (SAT practice test 1 article 3)

Correct answer: 
the deoxyribonucleic acid molecule.

Distractors:
\begin{enumerate}
\item the deoxyribonucleic acid coenzyme.
\item the deoxyribonucleic acid polymer.
\item the deoxyribonucleic acid trimer
\end{enumerate}

\begin{comment}
Question: 
What is a very long chain, the backbone of which consists of a regular alternation of sugar and phosphate groups? (SAT practice test 1 article 3)

Correct answer: 
The molecule.

Distractors:
\begin{enumerate}
\item The coenzyme.
\item The polymer.
\item The trimer.
\end{enumerate}
\end{comment}

\subsubsection*{Example 5}

Question: 
When does Deep Space Industries of Virginia hope to be harvesting metals from asteroids?
(SAT practice test 1 article 5)

Correct answer: 
by 2020.

Distractors:
\begin{enumerate}
\item by 2021.
\item by 2030.
\item by 2019.
\end{enumerate}

%The following is an example with one distractor without an adequate distracting effect.

\subsubsection*{Example 6}

%This example contains a distractor without distracting effect.

Question: 
What did a British study of the way women search for medical information online indicate?
(SAT practice test 2 article 3)

Correct answer: 
An experienced Internet user can, at least in some cases, assess the trustworthiness and probable value of a Web page in a matter of seconds.

Distractors:
\begin{enumerate}
\item An experienced Supernet user can, at least in some cases, assess the trustworthiness and probable value of a Web page in a matter of seconds.
\item An experienced CogNet user can, at least in some cases, assess the trustworthiness and probable value of a Web page in a matter of seconds.
\item An inexperienced Internet user can, at least in some cases, assess the trustworthiness and probable value of a Web page in a matter of seconds.
% (This distractor does not have
%a desired distracting effect as ``SUV user" is clearly not related to the context.)
\end{enumerate}

\subsubsection*{Example 7}

What does a woman know better than a man? (SAT test 2 article 4)

Correct answer: 
the cost of life.

Distractors:
\begin{enumerate}
\item the cost of happiness.
\item the cost of experience.
\item the risk of life.
\end{enumerate}

\subsubsection*{Example 8}

This example presents a distractor without sufficient distraction.

Question: 
What are subject to egocentrism, social projection, and multiple attribution errors?

Correct answer: 
their insights.

Distractors:
\begin{enumerate}
\item their perspectives.
\item their findings.
\item their valuables. 
\end{enumerate}
The last distractor can be spotted wrong by just looking at the question: It is easy to tell
that it is out of place without the need to read the article.

\section{Evaluations}

We implemented our method using the latest versions of POS tagger\footnote{https://nlp.stanford.edu/software/tagger.shtml}, NE tagger \cite{peters2017semi},
semantic-role labeling \cite{shi2019simple}, and fastText \cite{mikolov2018advances}.
We used the US SAT  practice reading tests\footnote{https://collegereadiness.collegeboard.org/sat/practice/full-length-practice-tests} as a dataset for evaluations. 
There are a total of eight SAT practice reading tests, each consisting of five articles for a total of 40 articles.
Each article in the SAT practice reading tests consists of around 25 sentences and we
generated about 10 QAPs from each article. To evaluate our distractor generation algorithm, we selected independently at random slightly over 100 QAPs. After removing a smaller number of QAPs with
pronouns as target words, we have a total of 101 QAPs for evaluations.

We generated 3 distractors for each QAP for a total of 303 distractors, and evaluated distractors 
based on the following criteria:
\begin{enumerate}
\item A distractor is \textsl{adequate} if it is grammatically correct and relevant to the question with distracting effects.
\item An MCQ is \textsl{adequate} if each of the three distractors is adequate.
\item An MCQ is \textsl{acceptable} if one or two  distractors are adequate. 
%\item \textsl{unacceptable} if none of the distractors is adequate.
\end{enumerate}
We define two levels of distracting effects: (1) \textsl{sufficient distraction}: It requires 
an understanding of the underlying article to choose the correct answer;
(2) \textsl{distraction}: It only requires an understanding of the underlying question to choose
the correct answer.
A distractor has no distracting effect if it can be determined wrong by just looking at
the distractor itself.

%
%as following three points:
%\begin{itemize}
%    \item The distractor is grammatically consistent with the context.
%    \item The distractor is relevant to the question with distracting effects.
%    \item The distractor provides enough confusion.
%\end{itemize}

\begin{comment}
\begin{table}[h]
\caption{Evaluation result}
\label{tab:evaluation-table} \centering
\begin{tabular}{|l|l|l|l|l|}
    \hline
                             & number of distractor & percentage of distractor & \\
    \hline                         
    Grammatically acceptable &   &   & \\
    \hline
    Relevantly acceptable    &   &   & \\
    \hline
    Confusion level          &   &   & \\
    \hline
\end{tabular}
\end{table}
\end{comment}

Evaluations were carried out by humans and the results are listed below:
\begin{enumerate}
\item All distractors generated by our method are grammatically correct.
\item 98\% distractors (296 out of 303) are relevant to the QAP with distraction.
\item 96\% distractors (291 out of 303) provide sufficient distraction.
\item 84\% MCQs are adequate.
\item All MCQs are acceptable (i.e., with at least one adequate distractor).
\end{enumerate}

\begin{comment}
0  grammatical issue
7  semantic issue
12  grammatical and semantic correct but no enough confusion

303 / 100% distractor is grammatically correct
296 / 98% distractor is relevant to the question with distracting effects
291 / 96% distractor provides enough confusion

16 / 16% MCQ acceptable if one or two distractors are adequate
0 MCQ unacceptable if none of the distractors is adequate.
\end{comment}

\chapter{Conclusion}

We presented 3 methods for generating MCQs.
\begin{enumerate}

\item  TP3, a deep-learning-based end-to-end Transformer with preprocessing and postprocessing pipelines, the large model achieved over 95\% of accuracy on Gaokao-EN dataset.

\item  MetaQA, a meta-sequence-learning-based scheme, which achieved 97\% of accuracy on SAT dataset.

\item  A distractor generation method, a combination of various NLP tools and algorithms, which also achieved a high accuracy.

\end{enumerate}

The TP3 works very well on unseen data, it can generate a large number of QAPs. 
On the other hand, the MetaQA can generate QAPs based on the meta sequences it has learned, but rarely generate inadequate question, which good complement the TP3.

All generated adequate questions are grammatically correct, and the corresponding answer is correct to the question.

The generated adequate distractors are the incorrect answers to the question, relate to the correct answer, and provide enough degree of distraction.

Overall, our methods for generating MCQs achieved over 95\% of accuracy.

 %Distractor Generation

\clearpage

\bibliographystyle{apalike}
\bibliography{bibliography}

\addcontentsline{toc}{chapter}{References}
% \include{bibliography}
% %\include{appendix-stub} 	%I used an appendix stub file before I wrote the real appendix so the main file would build correctly.
% \clearpage
% \addcontentsline{toc}{chapter}{Publications}
% \include{publication}
% \include{biography}
\appendix

\end{document}